\documentclass{article}

\usepackage{microtype}
\usepackage{graphicx}
\usepackage{subfigure}
\usepackage{booktabs} 

\usepackage{hyperref}

\usepackage[accepted]{icml2023}

\usepackage{amsmath}
\usepackage{amssymb}
\usepackage{mathtools}
\usepackage{amsthm}

\usepackage[capitalize,noabbrev]{cleveref}

\usepackage{hyperref}
\usepackage{url}
\usepackage{graphicx}
\usepackage{acronym}  
\usepackage{multirow} 
\usepackage{rotating} 
\usepackage{colortbl} 
\usepackage{multicol}

\newcommand\RotText[1]{\rotatebox{90}{\parbox{1.2cm}{\centering#1}}}

\usepackage[hang]{footmisc}
\theoremstyle{plain}

\theoremstyle{definition}

\theoremstyle{remark}

\usepackage[textsize=tiny]{todonotes}

\icmltitlerunning{QDT: Leveraging Dynamic Programming for Conditional Sequence Modelling in Offline RL}

\begin{document}

\twocolumn[
\icmltitle{Q-learning Decision Transformer: Leveraging Dynamic Programming for Conditional Sequence Modelling in Offline RL}



\icmlsetsymbol{equal}{*}

\begin{icmlauthorlist}
\icmlauthor{Taku Yamagata}{uob1}
\icmlauthor{Ahmed Khalil}{uob1}
\icmlauthor{Ra{\'u}l Santos-Rodr{\'i}guez}{uob1}
\end{icmlauthorlist}

\icmlaffiliation{uob1}{Intelligent System Laboratory, University of Bristol, Bristol, UK}

\icmlcorrespondingauthor{Taku Yamagata}{taku.yamagata@bristol.ac.uk}

\icmlkeywords{Reinforcement Learning, Dynamic Programming, Offline Reinforcement Learning, Decision Transformer, Machine Learning, ICML}

\vskip 0.3in
]



\printAffiliationsAndNotice{}

\begin{abstract}
Recent works have shown that tackling offline reinforcement learning (RL) with a conditional policy produces promising results. The Decision Transformer (DT) combines the conditional policy approach and a transformer architecture, showing competitive performance against several benchmarks. However, DT lacks stitching ability -- one of the critical abilities for offline RL to learn the optimal policy from sub-optimal trajectories. This issue becomes particularly significant when the offline dataset only contains sub-optimal trajectories.
On the other hand, the conventional RL approaches based on Dynamic Programming (such as Q-learning) do not have the same limitation; however, they suffer from unstable learning behaviours, especially when they rely on function approximation in an off-policy learning setting. In this paper, we propose the Q-learning Decision Transformer (QDT) to address the shortcomings of DT by leveraging the benefits of Dynamic Programming (Q-learning). It utilises the Dynamic Programming results to relabel the return-to-go in the training data to then train the DT with the relabelled data. Our approach efficiently exploits the benefits of these two approaches and compensates for each other's shortcomings to achieve better performance. 
\end{abstract}

\section{Introduction}
\label{sec:introduction}

The transformer architecture employs a self-attention mechanism to extract relevant information from high-dimensional data. It achieves state-of-the-art performance in a variety of applications, including natural language processing (NLP)~\citep{vaswani2017attention,radford2018improving,devlin2018bert} or computer vision~\citep{ramesh2021zero}. 
Its translation to the RL domain, the \Ac{DT}~\citep{chen2021decision}, successfully applies the transformer architecture to offline reinforcement learning tasks with good performance when shifting their focus on the sequential modelling. 
It employs a goal conditioned policy which converts offline RL into a supervised learning task, and it avoids the stability issues related to bootstrapping for the long term credit assignment~\citep{srivastava2019training,kumar2019reward,ghosh2019learning}. More specifically, \ac{DT} considers a sum of the future rewards --
\ac{RTG}, as the goal and learns a policy conditioned on the \ac{RTG} and the state. It is categorised as a \textit{reward conditioning} approach. Although \ac{DT} shows very competitive performance in the offline \ac{RL} tasks, it fails to achieve one of the desired properties of offline \ac{RL} agents, stitching. 
This property is an ability to combine parts of sub-optimal trajectories and produce an optimal one~\citep{fu2020d4rl}. 
We show a simple example of how \ac{DT} (\textit{reward conditioning} approaches) would fail to find the optimal path.
\begin{figure*}[t]
\centering
\hspace*{1.5cm}\includegraphics[width=0.8\textwidth]{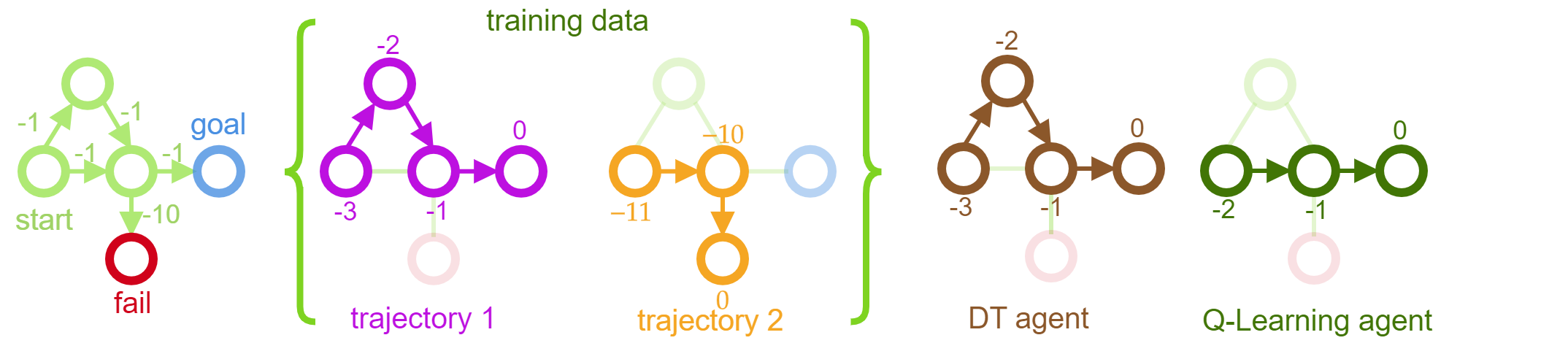}
\caption{A simple example demonstrates the decision transformer approach's issue (lack of \textit{stitching} ability) -- fails to find the shortest path to the goal. In contrast, Q-learning finds the shortest path. The numbers on the arrows are rewards on the path and the numbers on the states are \acp{RTG}.}
\label{fig:simple_example_1}
\end{figure*}
\begin{figure*}[t]
    \centering
    \includegraphics[width=0.85\textwidth]{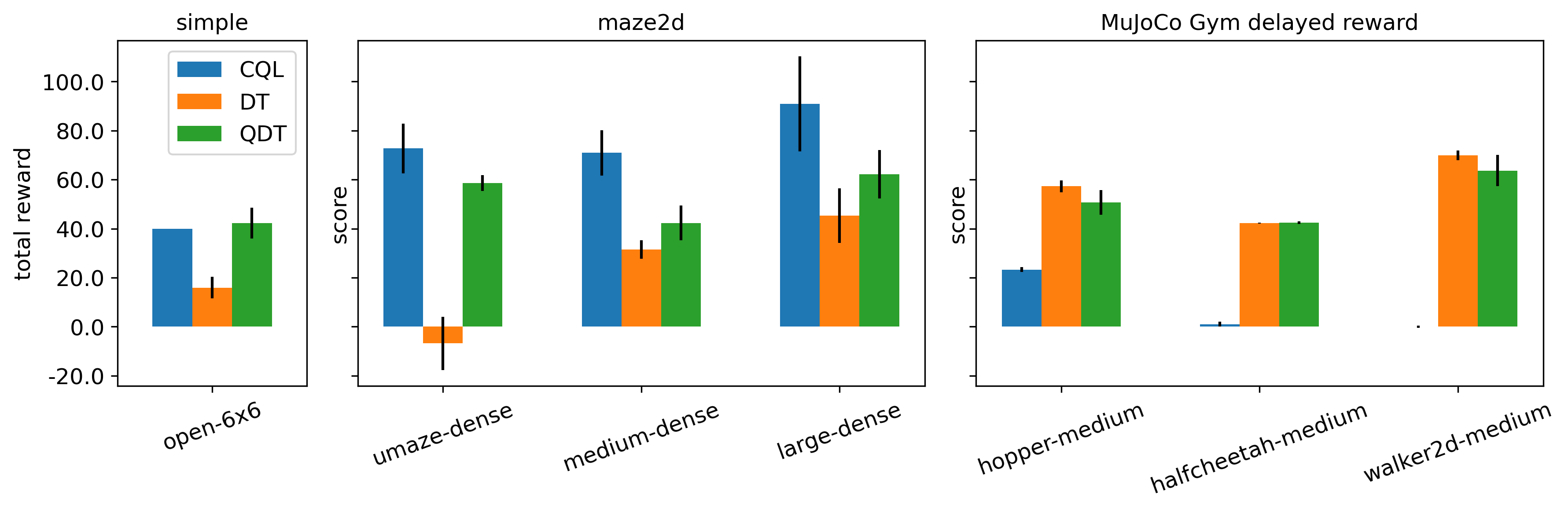}
    \caption{Evaluation results for conservative Q-learning (CQL), Decision Transformer (DT) and Q-learning Decision Transformer (QDT). The left two plots (simple and maze2d environments) show that the DT does not perform as it fails to stitch trajectories, and the right plot shows that CQL fails to learn from a sparse reward scenario (delayed reward). In contrast, QDT achieves consistently good results across all the environments.}
    \label{fig:results-preview}
\end{figure*}
To demonstrate the limitation of the reward conditioning approaches (\ac{DT}), consider a task to find the shortest path from the left-most state to the rightmost state without going down to the fail state in Fig.~\ref{fig:simple_example_1}. We set the reward as $-1$ at every time step and $-10$ for the action going down to the fail state.
The training data covers the optimal path, but none of the training data trajectories has the entire optimal path. The agent needs to combine these two trajectories and come up with the optimal path. 
The reward conditioning approach essentially finds a trajectory from the training data that gives the ideal reward and takes the same action as the trajectory. 
In this simple example, trajectory 2 has a meagre reward. 
Hence, it always follows the path of trajectory 1 despite trajectory 2 giving the optimal path for the first action. 

In contrast to the reward conditioning approaches (\ac{DT}), Q-learning\footnote{In this paper, we will use the \textit{Q-learning} and \textit{Dynamic Programming} interchangeably to indicate any \ac{RL} algorithm relying on the Bellman-backup operation.} does not suffer from the issue and finds the optimal path quickly in this simple example. 
Q-learning takes each time step separately and propagates the best future rewards backwards. Hence it can learn from the first optimal action from trajectory 2.
However, Q-learning has some issues on a long time horizon and sparse reward scenarios. 
It attempts propagating the value function backwards to its initial state, often struggling to learn across long time horizons and sparse reward tasks. This is especially true when Q-learning uses function approximation in an off-policy setting as discussed in Sec. 11.3 in \citep{Sutton1998}.

Here, we devise a method to address the issues above by leveraging Q-learning to improve DT. 
Our approach differs from other offline \ac{RL} algorithms that often propose a new single architecture of the agent and achieves better performance. We propose a framework that improves the quality of the offline dataset and obtains better performance from the existing offline \ac{RL} algorithms.
Our approach exploits the Q-learning estimates to relabel the \ac{RTG} in the training data for the \ac{DT} agent. The motivation for this comes from the fact that Q-learning learns \ac{RTG} value for the optimal policy. This suggests that relabelling the \ac{RTG} in the training data with the learned \ac{RTG} should resolve the \ac{DT} stitching issue. However, Q-learning also struggles in situations where the states require a large time step backward propagation. In these cases, we argue that \ac{DT} will help as it estimates the sequence of states and actions without backward propagation.
Our proposal (QDT) exploits the strengths of each of the two different approaches to compensate for other's weaknesses and achieve a more robust performance. Our main evaluation results are summarised in Fig.~\ref{fig:results-preview}. The left two plots (simple and maze2d environments) show that DT does not perform well as it fails to stitch trajectories, while the right plot illustrates that CQL (Q-learning algorithm for offline reinforcement learning) fails to learn in a sparse reward scenario (delayed reward). These results indicate that neither of these approaches works well for all environments, and we might have abysmal results by selecting the wrong type of algorithms. In contrast, QDT performs consistently well across all environments and shows robustness against different environments.
Through our evaluations, we also find that some of the evaluation results in the prior works may not be directly comparable, and it causes some contradicting conclusions. We touch on the issue in Sec.~\ref{sec:discussion}.


\section{Preliminaries}
\label{sec:preliminaries}

{\bf Offline Reinforcement Learning.}
The goal of \ac{RL} is to learn a policy maximising the expected sum of rewards in a \ac{MDP}, which is a four-tuple $(S,A,p,r)$ where $S$ is a set of states, $A$ is a set of actions, $p$ is the state transition probabilities, and $r$ is the reward.

In the online or on-policy \ac{RL} settings, an agent has access to the target environment and collects a new set of trajectories every time it updates its policy. The trajectory consists of $\{s_t, a_t, r_t\}_{t=0}^{T}$ where $s_t$, $a_t$ and $r_t$ are the state, action and reward at time $t$ respectively, and $T$ is the episode time horizon. In off-policy \ac{RL} case, the agent also has access to the environment to collect trajectories, but it can update its policy with the trajectories collected with other policies. Hence, it improves its sample efficiency as it can still make use of past trajectories. Offline \ac{RL} goes one step further than off-policy \ac{RL}. It learns its policy purely from a static dataset that is previously collected with an unknown behaviour policy (or policies). This paradigm can be precious in case of the interaction with the environment being expensive or high risk (e.g., safety critical applications).

{\bf Decision Transformers.}
\ac{DT} architecture~\citep{chen2021decision} casts the \ac{RL} problem as conditional sequence modelling. Unlike the majority of prior \ac{RL} approaches that estimate value functions or compute policy gradients, \ac{DT} outputs desired future actions from the target sum of future rewards \acp{RTG}, past states and actions.
\begin{multline}
\label{eq:dt_input}
    \tau = 
    (R_{t-K+1}, s_{t-K+1}, a_{t-K+1}, \cdots, \\
    R_{t-1}, s_{t-1}, a_{t-1}, R_t, s_t).
\end{multline}
Eq.~\ref{eq:dt_input} shows the input of a \ac{DT}, where $K$ is the context length, $R$ is \acp{RTG} ($R_t = \sum_{t'=t}^T r_{t'}$), $s$ is states and $a$ is actions. Then \ac{DT} outputs the next action ($a_t$). \ac{DT} employs Transformer architecture~\citep{vaswani2017attention}, which consists of stacked self-attention layers with residual connections. It has been shown that the Transformer architecture successfully relates scattered information in long input sequences and produces accurate outputs~\citep{vaswani2017attention,radford2018improving,devlin2018bert,ramesh2021zero}. 

{\bf Conservative Q learning.}
In this work, we use the \ac{CQL} framework~\citep{kumar2020conservative} for the Q-learning algorithm. \ac{CQL} is an offline \ac{RL} framework that learns Q-functions that are lower-bounds of the true values. It augments the standard Bellman error objective with a regulariser which reduces the value function for the out-of-distribution state-action pair while maintaining ones for state-action pairs in the distribution of the training dataset. In practice, it uses the following iterative update equation to learn the Q-function under a learning policy $\mu(a|s)$.
\begin{multline}
\label{eq:cql_update}
\hat{Q}^{k+1} \leftarrow \\
\arg\min_{Q} \alpha \left(\mathbb{E}_{s\sim\mathcal{D},a\sim\mu(a|s)}[Q(s,a)] - \mathbb{E}_{s,a\sim\mathcal{D}}[Q(s,a)]\right) \\
+ \frac{1}{2} \mathbb{E}_{
\footnotesize
\begin{matrix}
s,a,s'\sim\mathcal{D} \\
a'\sim\mu(a'|s')
\end{matrix}}\left[\left(r(s,a) + \gamma \hat{Q}^{k}(s',a') - Q(s,a)\right)^2\right],
\end{multline}
where $\mathcal{D}$ is the training dataset and $\gamma$ is a discount factor. \citet{kumar2020conservative} showed that while the resulting Q-function, $\hat{Q}^\mu := \lim_{k\rightarrow\infty}\hat{Q}^k$ may not be a point-wise lower-bound, it is a lower bound of $V(s)$, i.e. $\mathbb{E}_{\mu(a|s)}[\hat{Q}^\mu(s,a)] \leq V^\mu(s)$.

\section{Method}
\label{sec:method}
We propose a method that leverages Dynamic Programming approach (Q-learning) to compensate for the shortcomings of the reward conditioning approach (\ac{DT}) and build a robust algorithm for the offline RL setting.
Our proposal consists of three steps. First, the value function is learned with Q-learning. Second, the offline \ac{RL} dataset is refined by relabelling the \ac{RTG} values with the result of Q-learning. Finally, the \ac{DT} is trained with the relabelled dataset. The first and third steps do not require any modifications of the existing algorithms.

The reward conditioning approach (\ac{DT}) takes an entire trajectory sequence and conditions on it using the sum of the rewards for that given sequence. Such an approach struggles on tasks requiring \textit{stitching}~\citep{fu2020d4rl} -- the ability to learn an optimal policy from sub-optimal trajectories by combining them.
In contrast, the Q-learning alternative propagates the value function backwards for each time step separately with the Bellman backup, and pools the information for each state across trajectories. It therefore does not have the same issue.
Our proposal tackles the \textit{stitching} issue of the reward conditioning approach by relabelling the \ac{RTG} values with the learned Q-functions. With the relabelled dataset, the reward conditioning approach (\ac{DT}) can then utilize optimal sub-trajectories from their respective sub-optimal trajectories.

We now discuss how to relabel the \acp{RTG} values with the learned Q-functions. Replacing all of the \acp{RTG} values with Q-functions is not adequate because not all the learned Q-functions are accurate, especially in a long time horizon and sparse reward case. Ideally, we would like to replace the \acp{RTG} values where the learned Q-functions are accurate. 

In this work, we employ the \ac{CQL} framework for the offline Q-learning algorithm, which learns the lower bound of the value function. We replace the \acp{RTG} values when the \ac{RTG} in the trajectory is lower than the lower bound. With this approach, our method substitutes the \acp{RTG} values where the learned value function is indeed accurate (or closer to the true values). We also replace all \ac{RTG} values prior to the replaced \ac{RTG} along with the trajectory by using reward recursion ($R_{t-1} = r_{t-1} + R_t$). This propagates the replaced \ac{RTG} values to all the time steps prior to the replaced point. 

To apply this idea, we initialise the last state \ac{RTG} to zero ($R_T=0$), and then we start the following process from the end of the trajectory to the initial state backwards in time. First, the state value is computed for the current state with the learned value function $\hat{V}(s_{t}) = \mathbb{E}_{a\sim\pi(a|s_{t})}[\hat{Q}(s_{t},a)]$,
where the $\pi$ is the learned policy. Next, the value function is compared ($\hat{V}(s_{t})$) against the \ac{RTG} value for the current state ($R_{t}$). If the value function is greater than that of the \ac{RTG}, the \ac{RTG} for the previous time step is set from ($R_{t-1}$) to $r_{t-1} + \hat{V}(s_{t})$, otherwise it is set to $r_{t-1} + R_{t}$. We repeat this process until the initial state is reached. This process is summarised in Algorithm~\ref{alg:replace_R}. 

The above relabelling process might introduce inconsistencies between the reward and \ac{RTG} within the \ac{DT} input sequence (Eq.~\ref{eq:dt_input}). The \ac{RTG} value is sum of the future rewards, hence it must always be $R_{t} = r_t + R_{t+1}$. However, the relabelling process might break this relationship. To maintain this consistency within the input sequence of \ac{DT}, we regenerate the \ac{RTG} for the input sequence ($\{\hat{R}_{t-K+1}, \cdots, \hat{R}_{t-1}, \hat{R}_t\}$) by copying the last \ac{RTG} ($\hat{R}_t=R_t$) and then repeatedly apply $\hat{R}_{t'} = r_t' + \hat{R}_{t'+1}$ backwards until $t'=t-K+1$. We repeat this for each the input sequences to maintain the consistency of the rewards and \acp{RTG}. This process is summarised in Algorithm~\ref{alg:replace_R_2}.

\begin{algorithm}
\caption{Relabelling return-to-go}\label{alg:replace_R}
\begin{algorithmic}
\STATE {\bfseries Input:} rewards $r_{1:T}$, learned value function $\hat{V}(s)$, trajectory length $T$
\STATE {\bfseries Output:} relabelled return to go $R_{1:T}$
\STATE $R_T \gets 0$
\STATE $\tau \gets T$
\WHILE{$\tau > 0$}
    \STATE $R_{\tau-1} \gets r_{\tau-1} + \max(R_{\tau}, \hat{V}(s_{\tau}))$
    \STATE $\tau \gets \tau - 1$
\ENDWHILE
\end{algorithmic}
\end{algorithm}
\begin{algorithm}
\caption{Generating return-to-go for DT}\label{alg:replace_R_2}
\begin{algorithmic}
\STATE {\bfseries Input:} rewards $r_{t-K+1:t}$, return to go for time $t$ $R_t$, context length $K$
\STATE {\bfseries Output:} relabelled return to go for DT $\hat{R}_{1:T}$
\STATE $\hat{R}_t \gets R_t$
\STATE $\tau \gets t-1$
\WHILE{$\tau > t-K$}
    \STATE $\hat{R}_\tau \gets r_\tau + \hat{R}_{\tau+1}$
    \STATE $\tau \gets \tau - 1$
\ENDWHILE
\end{algorithmic}
\end{algorithm}

{\bf Theoretical considerations of QDT.} 
\ac{QDT} relies on \ac{DT} as the agent algorithm, which can be seen as a reward conditioning model. A reward conditioning model takes the states and \ac{RTG} as inputs and outputs actions. If we assume the model is trained with the state $s_t$ and {the optimal action $a_t$ together with} the optimal state-action value function ($Q^*(s_t,a_t)$), then we can guarantee that the model will output the optimal action ($\arg\max_a Q^*(s_t,a)$) for as long as it is given $s_t$ and $\max_a Q^*(s_t,a)$ as inputs \citep{srivastava2019training}. In practice, we do not know the optimal value function $Q^*(s,a)$, hence \ac{DT} (and similarly other reward conditioning approaches) uses \ac{RTG} instead. \ac{RTG} is collected through the behaviour policy (or policies) and often is not optimal -- with the majority of values being much lower than the corresponding optimal value function ($Q^*(s,a)$). As \ac{QDT} uses \ac{CQL} to learn the optimal \textit{conservative} value function, Th.~3.2 in \citet{kumar2020conservative} shows that the \textit{conservative} value function is a lower bound of the true value function. Hence the \ac{QDT} relabelling process moves the \ac{RTG} in the training dataset closer to the optimal value function (see Appendix~\ref{apd:relabelling_rtg}).

\section{Related Work}
\label{sec:related_work}
{\bf Offline reinforcement learning.}
The offline \ac{RL} learns its policy purely from a static dataset that was previously collected with an unknown behaviour policy (or policies). As the learned policy might differ from the behaviour policy, the offline algorithms must mitigate the effect of the \textit{distributional shift}~\citep{agarwal2020optimistic,prudencio2022survey}. One of the most straightforward approaches to address the issue is by constraining the learned policy to the behaviour policy~\citep{fujimoto2019off,kumar2019stabilizing,wu2019behavior}. Other methods constrain the learned policy by making conservative estimates of future rewards~\citep{kumar2020conservative,yu2021combo,fujimoto2021minimalist}. Some model-based methods estimate the model's uncertainty and penalize the actions whose consequences are highly uncertain~\citep{janner2019trust,kidambi2020morel}. Some approaches address the \textit{distributional shift} without restricting the learned policy. One such approach group is weighted imitation learning~\citep{Wang2018marwil,Peng2019awr,Wang2020crr,Nair2020awac,Chen2020bail,Siegel2020,Brandfonbrener2021}, which carries out imitation learning by putting higher weights on the good state-action pairs. It usually uses an estimated advantage function as the weight. As this approach imitates the selected parts of the behaviour policy, and it naturally restricts the learned policy within the behaviour policy.
The other group of the approaches without restricting the learning policy is conditional sequence modelling, which learns a policy conditioned with a particular metric for the future trajectories. Some examples of the metrics are sum of the future rewards~\citep{srivastava2019training,chen2021decision}, a certain state (sub goal)~\citep{codevilla2018end,ghosh2019learning,lynch2020learning} and even learned features from the future trajectory~\citep{furuta2021generalized}.

Our approach does not belong to any of these groups but is related to the approach of learning pessimistic value function, the conditional sequence modelling and weighted imitation learning approaches. Essentially, our method is a conditional sequence modelling approach as it learns the following action conditioned on the current state and the sum of the future rewards, but the training data is augmented by the result of the learned pessimistic value function. Also, the overall high-level structure is somewhat similar to the weighted imitation learning, which learns the value function and uses it to weight the training data in the following imitation learning stage. However, each component is very different from ours, and it uses the value function to weight the training data, whereas our approach relabels the \ac{RTG} values by tracing back the trajectory with the learned value function as well as the trajectory itself where the learned value function is not reliable. Also, in our approach, the policy is learned with conditional sequence modelling, whereas they use non-conditional non-sequential models. We can apply our relabelling approach to the weighted imitation learning algorithms, which is an exciting future avenue.

{\bf Data centric approach.}
Andrew Ng recently spoke about the importance of the training data to achieve good performance from a machine learning model and suggests we should spend more of our effort on data than on the model (Data-centric Approach)~\citep{press2021datacentric}. He said, "In the \textit{Data-centric Approach}, the consistency of the data is paramount and using tools to improve the data quality that will allow multiple existing models to do well." Our method can be seen as \textit{Data-centric Approach} for offline RL, as we focus on improving the training data and using the existing models. Our method provides a tool to improve data quality.


\section{Evaluation}
\label{sec:evaluation}
We investigate the performance of \ac{QDT} relative to the offline \ac{RL} algorithm with the Dynamic Programming based approach as well as the reward conditioning approach. As \ac{QDT} utilises the result of \ac{CQL} and it is considered as the state-of-art offline RL method, we pick \ac{CQL} as the benchmark for the Dynamic Programming based approach and \ac{DT} for the reward conditioning approach for the same reason.
From the evaluations in this section, we would like to demonstrate the benefits and weaknesses of the Dynamic Programming approach and reward conditioning approach and how (\ac{QDT}) mitigates their weaknesses. We start our investigation with a simple environment with sub-optimal trajectories. As it is a simple environment,  a Dynamic Programming approach (\ac{CQL}) should work well, and as it uses sub-optimal trajectories, the reward conditioning approach (\ac{DT}) will struggle. It is interesting to see how much \ac{QDT} helps in the circumstance.
We also evaluate them on Maze2D environments designed to test the stitching ability with different levels of complexity. We expect that \ac{DT} struggle whereas \ac{CQL} and \ac{QDT} performs well on them.
Then, we evaluate the algorithms on complex control tasks -- Open AI Gym MuJoCo environments with delayed (sparse) reward as per \citet{chen2021decision}. They have zero rewards at all the non-terminal states and put the total reward at the terminal state. It should make the Dynamic Programming approach (\ac{CQL}) learning harder as it requires propagating the reward from the terminal state all way to the initial state.
Finally, we show the evaluation results for Open AI Gym MuJoCo environments with the original dense reward setting for the reference. 

{\bf Simple environment.}
To highlight the benefit of QDT, we evaluate our method in a simple environment, which has 6-by-6 discrete states and eight discrete actions. The goal of the task is to find the shortest path from the start to the goal state. We prepare an offline \ac{RL} dataset with a hundred episodes from a uniformly random policy and then remove an episode that achieves close to the optimal total reward to make sure it only contains sub-optimal trajectories. Refer to Appendix~\ref{apd:simple_env_eval} for details of the environment and dataset.

\vspace{-0.5cm}
\begin{table}[h]
  \caption{Simple Environment Evaluation Results. Average and standard deviation scores are reported over 10 seeds.}
  \label{tab:simple_env_result}
  \centering
  \begin{tabular}{lccc}
    \hline
        & CQL & DT & QDT \rule[0mm]{0mm}{3.2mm}\\
    \hline
    Total Reward & $\textbf{40.0} \pm 0.0$ & $15.9 \pm 4.4$ & $\textbf{42.2} \pm 6.3$\rule[0mm]{0mm}{3.5mm} \\
    \hline
  \end{tabular}
\end{table}

Table~\ref{tab:simple_env_result} shows the summary of the evaluation results. We also evaluate the performance of \ac{CQL}, which is used for relabeling. It shows vanilla \ac{DT} fails badly, which indicates \ac{DT} struggles to learn from sub-optimal trajectories, whereas \ac{CQL} performs well as it employs a Dynamic Programming approach, which can pool information across trajectories and successfully figure out the near-optimal policy. It shows \ac{QDT} performs similar to \ac{CQL}, which indicates that although \ac{QDT} uses the conditional policy approach, it overcomes its limitation and learns the near-optimal policy from the sub-optimal data. Further details in Appendix~\ref{apd:simple_env_eval}.

\begin{figure}
    \centering
    \includegraphics[width=0.9\columnwidth]{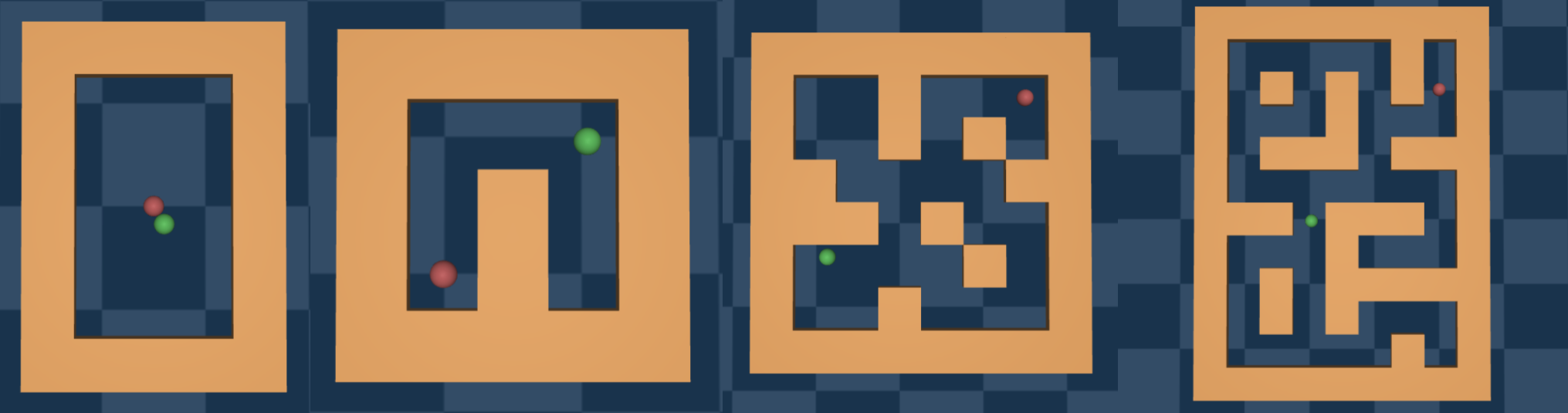}
    \caption{Four Maze2D environment layouts (from left to right: open, umaze, medium and large).}
    \label{fig:maze2d-view}
\end{figure}

{\bf Maze2D environments.}
Maze2D domain is a navigation task requiring an agent to reach a fixed goal location. The tasks are designed to provide tests of the ability of offline RL algorithms to be able to stitch together parts of different trajectories~\citep{fu2020d4rl}. It has four kinds of environments -- open, umaze, medium and large, and they are getting more complex mazes in the order (Fig.~\ref{fig:maze2d-view}) \footnote{https://github.com/rail-berkeley/d4rl/wiki/Tasks}. 
Also, it has two kinds of reward functions -- normal and dense. The normal gives a positive reward only when the agent reaches the goal, whereas the dense gives the rewards at every step exponentially proportional to the negative distance between the agent and the goal. 
For the model, we use the \ac{DT} source code provided by the authors \footnote{https://github.com/kzl/decision-transformer} and d3rlpy \footnote{https://github.com/takuseno/d3rlpy}~\citep{seno2021d3rlpy} -- offline RL library for \ac{CQL}, then build \ac{QDT} by replacing the return-to-go in the \ac{DT} before its training.
\begin{table*}[h]
\renewcommand{\arraystretch}{1.0}
  \centering
  \caption{Maze2D Evaluation Results. Average and standard deviation scores are reported over 5 seeds. The result for each seed is obtained by evaluating the last learned model on the target environment. The best average values are marked in bold.}
  \label{tab:maze2d_results}
  \begin{tabular}{>{\centering\arraybackslash}p{0.5cm}lcp{0mm}cp{0mm}c}
    \hline
       & Dataset & CQL & & DT & & QDT \\
    \hline
    \multirow{4}*{\RotText{Sparse Reward}}
        & maze2d-open-v0        & $\mathbf{216.7}\pm80.7$ && $196.4\pm39.6$ && $190.1\pm37.8$\rule[0mm]{0mm}{3.mm}\\ 
        & maze2d-umaze-v1       & $\mathbf{94.7}\pm23.1$  && $31.0\pm21.3$  && $57.3\pm8.2$\rule[0mm]{0mm}{3.mm}\\
        & maze2d-medium-v1      & $\mathbf{41.8}\pm13.6$  && $8.2\pm4.4$   && $13.3\pm5.6$\rule[0mm]{0mm}{3.mm}\\
        & maze2d-large-v1       & $\mathbf{49.6}\pm8.4$   && $2.3\pm0.9$    && $31.0\pm19.8$\rule[0mm]{0mm}{3.mm}\\
    \hline
    \multirow{4}*{\RotText{Dense Reward}}
        & maze2d-open-dense-v0  & $307.6\pm43.5$ && $346.2\pm14.3$ && $\mathbf{325.7}\pm61.4$\rule[0mm]{0mm}{3.mm}\\
        & maze2d-umaze-dense-v1 & $\mathbf{72.7}\pm10.1$  && $-6.8\pm10.9$  && $58.6\pm3.3$\rule[0mm]{0mm}{3.mm}\\
        & maze2d-medium-dense-v1& $\mathbf{70.9}\pm9.2$   && $31.5\pm3.7$   && $42.3\pm7.1$\rule[0mm]{0mm}{3.mm}\\
        & maze2d-large-dense-v1 & $\mathbf{90.9}\pm19.4$  && $45.3\pm11.2$  && $62.2\pm9.9$\rule[0mm]{0mm}{3.mm}\\
    \hline
  \end{tabular}
\end{table*}
Table~\ref{tab:maze2d_results} shows the summary of the results. We report the normalised total reward (score) such that 100 represents an expert policy~\citep{fu2020d4rl}. \ac{CQL} works well, especially with the dense rewards. \ac{DT} struggles in many cases due to the lack of stitching ability. (These environments are designed to test the stitching ability.) \ac{QDT} clearly improves \ac{DT} performance, especially where \ac{CQL} works well. It indicates that \ac{QDT} brings the stitching capability to \ac{DT} approach. We discuss the performance gap between \ac{CQL} and \ac{QDT} in Sec.~\ref{sec:discussion}.

{\bf Open AI Gym MuJoCo environments with delayed (sparse) reward.}
We also evaluate our approach (\ac{QDT}) on complex control tasks -- Open AI gym MuJoCo environments with the D4RL offline RL datasets~\citep{fu2020d4rl}. 
The Open AI gym MuJoCo environments consist of three tasks \textit{Hopper}, \textit{HalfCheetah} and \textit{Walker2d}. We test on \textit{medium} and \textit{medium-replay} v2 datasets. 
To demonstrate the shortcoming of the Dynamic Programing approach (\ac{CQL}), we follow \citet{chen2021decision} and evaluate the algorithms with a delayed (sparse) reward scenario in which the agent does not receive any reward along the trajectory and receives the sum of the rewards at the final time step.
Again we use the \ac{DT} and \ac{CQL} models from the existing source code for the MuJoCo Gym environments without any modifications and add extra code for the relabelling of the \ac{RTG} values.
\begin{table*}[h]
\renewcommand{\arraystretch}{1.0}
  \centering
    \caption{Open AI Gym MuJoCo with Delayed Reward Evaluation Results. Average and standard deviation scores are reported over 5 seeds. Our simulation results are in the Results columns, best average boldfaced. Ref.$^{*2}$ are the results copied from \citet{chen2021decision}. We are not sure which version of dataset the authors used for Ref.$^{*2}$, and only \textit{Hopper} results are available in the paper.}
    \label{tab:gym_mujoco_sparse_results}
    \begin{tabular}{>{\centering\arraybackslash}p{0.5cm}
                  p{2.3cm}
                  >{\centering\arraybackslash}p{1.6cm}
                  >{\columncolor[rgb]{0.9, 0.9,0.9}\centering\arraybackslash}p{0.8cm}
                  p{0mm}
                  >{\centering\arraybackslash}p{1.7cm}
                  >{\columncolor[rgb]{0.9, 0.9,0.9}\centering\arraybackslash}p{1.5cm}
                  p{0mm}
                  >{\centering\arraybackslash}p{1.6cm}
                  }
    \hline
        &   & \multicolumn{2}{c}{CQL} & & \multicolumn{2}{c}{DT} & & QDT \\
    \multicolumn{2}{c}{Dataset}
            & Results& Ref.$^{*2}$    & & Results& Ref.$^{*2}$   & & Results \\
    \hline
    \multirow{3}*{\RotText{Medium}} 
        & Hopper-v2      & $23.3\pm1.0$ & $5.2$ && $\mathbf{57.3}\pm2.4$   & $60.7\pm4.5$ && $50.7\pm5.0$\rule[0mm]{0mm}{2.5mm} \\
        & HalfCheetah-v2 & $ 1.0\pm1.0$ & $-$   && $42.2\pm0.2$            & $-$          && $\mathbf{42.4}\pm0.5$\rule[0mm]{0mm}{2.5mm} \\
        & Walker2d-v2    & $ 0.0\pm0.4$ & $-$   && $\mathbf{69.9}\pm2.0$ & $-$          && $63.7\pm6.4$\rule[0mm]{0mm}{2.5mm} \\
    \hline
    \multirow{3}*{\RotText{Medium Replay}}
        & Hopper-v2      & $ 7.7\pm5.9$ & $2.0$ && $\mathbf{50.8}\pm14.3$ & $78.5\pm3.7$ && $38.7\pm26.7$\rule[0mm]{0mm}{2.5mm}\\
        & HalfCheetah-v2 & $ 7.8\pm6.9$ & $-$   && $\mathbf{33.0}\pm4.8$ & $-$           && $32.8\pm7.3$\rule[0mm]{0mm}{2.5mm}\\
        & Walker2d-v2    & $ 3.2\pm1.7$ & $-$   && $\mathbf{51.6}\pm24.6$ & $-$          && $29.6\pm15.5$\rule[0mm]{0mm}{2.5mm}\\
    \hline
  \end{tabular}
\end{table*}
Table~\ref{tab:gym_mujoco_sparse_results} shows the simulation results (scores) for the delayed reward case.  We also copy the simulation results from \citet{chen2021decision} for \ac{DT} and \ac{CQL} for the reference. All of the numbers in the table are the normalised total reward (score) such that 100 represents an expert policy~\citep{fu2020d4rl}. 
As expected, \ac{CQL} struggles to learn a good policy, whereas the \ac{DT} shows good performance. Also, \ac{QDT} performs similar to \ac{DT} even though they are using the results of \ac{CQL} that performs badly. It indicates that \ac{QDT} successfully use the information from \ac{CQL} where it is useful. One exception is the medium-replay-walker2d result. \ac{QDT} performs worse than \ac{DT} here. Through some investigations, we found that the \ac{CQL} algorithm overestimates the value function in the majority of the states in the medium-replay-walker2d dataset. We touch the issue in the discussion section.

{\bf Open AI Gym MuJoCo environments.}
We also evaluate our approach (\ac{QDT}) on Open AI gym MuJoCo environments with the original dense reward for the reference. As they have dense rewards and contain reasonably good trajectories, both \ac{CQL} and \ac{DT} would work well. 
\begin{table*}[h]
\renewcommand{\arraystretch}{1.0}
  \centering
  \caption{Open AI Gym MuJoCo Evaluation Results. Average and standard deviation scores are reported over 5 seeds. Our simulation results are in Results columns. The best average values are marked in bold. Ref.$^{*1}$ is the results copied from \citet{emmons2021rvs}. Ref.$^{*2}$ is the results copied from \citet{chen2021decision}.}
  \label{tab:gym_mujoco_result}
  \begin{tabular}{>{\centering\arraybackslash}p{0.5cm}
                  p{2.3cm}
                  >{\centering\arraybackslash}p{1.7cm}
                  >{\columncolor[rgb]{0.9, 0.9,0.9}\centering\arraybackslash}p{0.8cm}
                  p{0mm}
                  >{\centering\arraybackslash}p{1.6cm}
                  >{\columncolor[rgb]{0.9, 0.9,0.9}\centering\arraybackslash}p{1.6cm}
                  p{0mm}
                  >{\centering\arraybackslash}p{1.6cm}
                  }
    \hline
        &   & \multicolumn{2}{c}{CQL} & & \multicolumn{2}{c}{DT} & & QDT \\
    \multicolumn{2}{c}{Dataset}
            & Results& Ref.$^{*1}$    & & Results& Ref.$^{*2}$   & & Results \\
    \hline
    \multirow{3}*{\RotText{Medium}} 
        & Hopper-v2      & $\mathbf{69.4}\pm13.1$ & $64.6$ && $60.3\pm5.5$ & $67.6\pm1.0$ && $66.5\pm6.3$\rule[0mm]{0mm}{3.2mm} \\
        & HalfCheetah-v2 & $\mathbf{49.2}\pm0.5$  & $49.1$ && $42.1\pm0.5$ & $42.1\pm0.1$ && $42.3\pm0.4$\rule[0mm]{0mm}{3.2mm} \\
        & Walker2d-v2    & $\mathbf{83.0}\pm0.6$  & $82.9$ && $73.3\pm2.5$ & $74.0\pm1.4$ && $67.1\pm3.2$\rule[0mm]{0mm}{3.2mm} \\
    \hline
    \multirow{3}*{\RotText{Medium Replay}}
        & Hopper-v2      & $\mathbf{96.2}\pm7.9$  & $97.8$ && $63.7\pm12.2$ & $82.7\pm7.0$ && $52.1\pm20.3$\rule[0mm]{0mm}{3.2mm} \\
        & HalfCheetah-v2 & $\mathbf{49.8}\pm0.5$  & $47.3$ && $34.1\pm1.1$  & $36.6\pm0.8$ && $35.6\pm0.5$\rule[0mm]{0mm}{3.2mm} \\
        & Walker2d-v2    & $\mathbf{76.5}\pm21.1$ & $86.1$ && $60.2\pm13.9$ & $66.6\pm3.0$ && $58.2\pm5.1$\rule[0mm]{0mm}{3.2mm} \\
    \hline
  \end{tabular}
\end{table*}
Table~\ref{tab:gym_mujoco_result} shows the summary of our simulation results for \ac{CQL}, \ac{DT} and \ac{QDT}. We also copy the simulation results from \citet{chen2021decision} for \ac{DT} and \citet{emmons2021rvs} for \ac{CQL} for the reference. 
Firstly, we can see that our simulation results are aligned with the references except for the medium-replay-hopper result. Because it has a relatively high variance, it is probably due to the small number of samples (five random seeds). Secondly, \ac{CQL} performs equal or better than \ac{DT} and \ac{QDT} in this evaluation. It is understandable as they have dense rewards (they do not require propagating value function in the trajectory). 
Finally, from the comparison between \ac{DT} and \ac{QDT}, \ac{QDT} performs the same as \ac{DT}. 

\section{Discussion}
\label{sec:discussion}

{
Our experiments show that \ac{QDT} is the only algorithm performing well across all environments. Although \ac{CQL} is very successful in many environments, it completely fails in the delayed reward MuJoCo cases. Similarly, \ac{DT} performs well on the delayed reward MuJoCo environments and fails in Maze2D environments. QDT, on the other hand, shows a higher level of robustness across types of environments.
This Section elaborates and reflects on the most relevant findings and properties of the \ac{QDT} framework. 
}

\begin{figure}
\centering
\includegraphics[width=0.98\columnwidth]{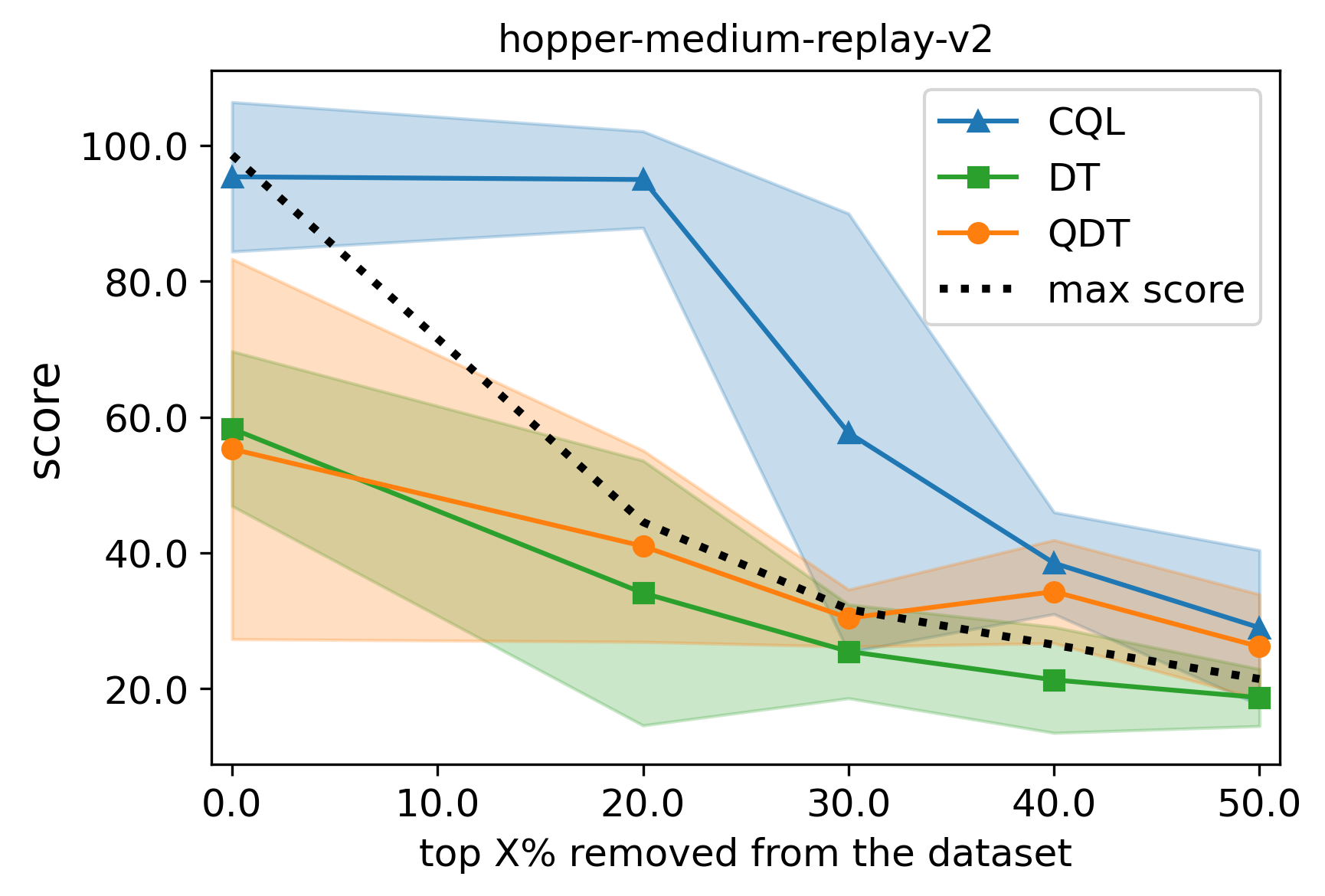}
\caption{Results for hopper-medium-replay-v2 dataset removed top X\% trajectories with one standard deviation. Maximum score in the dataset as a reference. CQL results are generally better than the maximum score, which indicates CQL successfully stitches sub-optimal trajectories, whereas DT fails to do so. QDT improves DT through relabelling, being better than the maximum score on the right-hand side of the plot.
}\label{fig:x-pct-hopper}
\end{figure}

{\bf Stitching ability.}
To demonstrate the stitching ability, we evaluate the performance of each algorithm with varying degrees of the sub-optimal dataset. We pick the medium-replay dataset for the MuJoCo Gym environment as it contains trajectories generated by various agent levels and removes the best $X$\% of the trajectories. As $X$ is increased, more good trajectories are removed from the dataset. Thereby moving further away from the optimal setup. Fig.~\ref{fig:x-pct-hopper} shows the \ac{CQL}, \ac{DT} and \ac{QDT} results as well as the best trajectory return in the dataset. It shows that \ac{CQL} offers better results than the best trajectory within the dataset except $X=0$, where the trajectory contains the best score; hence it can not be better than that. In contrast, \ac{DT} fails to exceed the best trajectory, which indicates \ac{DT} fails to stitch the sub-optimal trajectories. \ac{QDT} performs better than DT and becomes close to the CQL results at $X=40$ and $50$ (in the regime with $60-50\%$ bottom trajectories).

{\bf Performance gap between QDT and CQL.}
Although \ac{QDT} improves \ac{DT} on the sub-optimal dataset scenario (Table~\ref{tab:maze2d_results}), \ac{QDT} does not perform as well as \ac{CQL}. The results from \citet{emmons2021rvs} indicate that \ac{DT} can perform as well as \ac{CQL} when plenty of good trajectories are available (medium-expert dataset). It implies that there is still room for improvements for \ac{DT} and \ac{QDT} approaches with datasets that contain far from optimal trajectories. As a matter of fact, our experiment in the following subsection shows \ac{QDT} can perform as well as \ac{CQL} on the Maze2D environments in a certain condition. We believe it shows the \ac{QDT} approach has a good potential for further improvements.

{\bf The role of the discount factor.}
Our experiments use the discount factor $\gamma=0.99$ for \ac{CQL} as per the original paper. The relatively large discount (small $\gamma$ value) helps \ac{CQL} learning stability, and the value function estimation converges quickly to the correct value. Also, it makes the value function learn the discounted value -- the value function becomes less than the \ac{RTG} value, especially where the states are far from those giving positive rewards. The discounted value function can both positively and negatively affect the performance of the \ac{QDT}. Because the \acp{RTG} relabelling happens only when the value function gives a higher value than the \ac{RTG}, the discounted value function is less likely to be used. As a result, \ac{QDT} may fail to exploit all information from Q-learning (\ac{CQL}). Introducing the discount factor into the \ac{RTG} computation is the most straightforward approach to prevent this effect, by using Eq.~\ref{eq:dicounted_relabelling} in the relabelling, 
\begin{equation}
\label{eq:dicounted_relabelling}
\begin{split}
R_{\tau-1} &\gets r_{\tau-1} + \gamma \max(R_{\tau}, \hat{V}(s_{\tau})) \\
\hat{R}_\tau &\gets r_\tau + \gamma \hat{R}_{\tau+1}
\end{split}
\end{equation}
To confirm this idea, we evaluated \ac{QDT} with the discounted relabelling on maze2d-umaze-v1, -medium-v1 and -large-v1 datasets. The results  (Table~\ref{tab:maze2d_results_discount_rtg}) are with $\gamma=0.99$ for maze2d-umaze-v1 and -medium-v1 dataset and $\gamma=0.999$ for maze2d-large-v1 dataset. In this way, all of them achieve the performance of \ac{CQL}.
\begin{table*}[h]
\renewcommand{\arraystretch}{1.0}
  \centering
  \caption{Experimental Maze2D Evaluation Results. QDT with discounted RTGs results are obtained with $\gamma=0.99$ for maze2d-umaze and -medium, $\gamma=0.999$ for maze2d-large. The best average values are marked in bold.}
  \label{tab:maze2d_results_discount_rtg}
  \begin{tabular}{lcp{0mm}cp{0mm}cp{0mm}c}
    \hline
        Dataset & CQL & & DT & & QDT & & QDT with discounted RTGs\rule[0mm]{0mm}{4mm}\\
    \hline
        maze2d-umaze-v1       & $\mathbf{94.7}\pm23.1$  && $31.0\pm21.3$  && $57.3\pm8.2$ && $82.9\pm8.8$\rule[0mm]{0mm}{4mm}\\
        maze2d-medium-v1      & $41.8\pm13.6$  && $8.2\pm4.4$   && $13.3\pm5.6$ && $\mathbf{48.5} \pm 9.4$\rule[0mm]{0mm}{3.5mm}\\
        maze2d-large-v1       & $49.6\pm8.4$   && $2.3\pm0.9$   && $31.0\pm19.8$ && $\mathbf{62.0} \pm 13.4$\rule[0mm]{0mm}{3.5mm}\\
    \hline
  \end{tabular}
\end{table*}
Although we can improve \ac{QDT} results, this requires using different discount factors. We observe that a small $\gamma$ helps the convergence of Q-learning as it shortens the time horizon to consider. However, it becomes a disadvantage for long-time horizon environments. 
On the other hand, a large $\gamma$ helps maintain the rewards in a long-time horizon. 
However, this makes the convergence of  Q-learning harder. Hence, a small $\gamma$ (0.99) works well for short-time horizon tasks (umaze and medium), while a large $\gamma$ (0.999) is suitable for long-time horizon tasks (large). 
To resolve this issue, we need modifications to the existing (or an altogegher new) Q-learning (\ac{CQL}) algorithm. 
Using a value function for the \ac{QDT} relabelling requires an accurate estimate of values. However, traditional Q-learning algorithms tend to prioritise the advantage function, which represents the difference between the values of different actions over the actual value in the value function. It is crucial to employ a Q-learning algorithm that prioritises accurate value function estimation, as it will enhance the effectiveness of \ac{QDT} by providing more precise values for the relabeling process.


{\bf Conservative weight.}
\ac{CQL} has a hyperparameter called \textit{conservative weight}, denoted by $\alpha$ in Eq.~\ref{eq:cql_update}. It weights the regulariser term, where the higher value, the more conservative are the value function estimations. Ideally, we would like to set it as small as possible so that the estimated value function becomes a tighter lower bound; however, too small conservative weight might break the lower bound guarantee, and the learned value function might give a higher value than the true value~\citep{kumar2021workflow}. Empirically, we discovered that this is exactly what happens in our delayed reward experiment (Table~\ref{tab:gym_mujoco_sparse_results}) for the medium-replay-waker2d dataset example. The  value function learned by \ac{CQL} in the dataset has higher values than the corresponding true value in many states, and it causes the wrong relabelling of \ac{RTG} and, subsequently, a worse \ac{QDT} performance. We evaluated it with higher $\alpha$ values -- increased from 5.0 to 100. Though this improves the \ac{QDT} result from $29.6\pm15.5$ to $46.9\pm13.8$, it is still worse than \ac{DT}. This is left for future work for further investigation.
In this paper, we assume we have access to the environment in order to optimise the hyperparameters. However, this should be done purely offline for a proper offline RL setting. Although there are some proposals~\citep{paine2020hyperparameter,fu2021benchmarks,emmons2021rvs}, this is still an active research area. 

{
{\bf Alternative approaches.} 
\ac{QDT} introduces the stitching capability to \ac{DT} and achieves competitive performance across various environments. Concurrently to our work, \citet{hepburn2022model} proposes a method to stitch trajectories using a learned state transition model and value function, and produces near the optimal trajectories. This is a model-based approach and requires a dedicated mechanism for the stitching operation, whereas ours is model-free and employs an existing well-studied method (e.g. Q-learning) to stitch the trajectories.

Another alternative to achieve robust performance across all environments is to improve Q-learning (Dynamic Programming) for delayed reward scenarios,  for example, using the CQL agent with eligibility traces~\citep{precup2000eligibility, geist2014off, daley2023trajectory}, multi-step temporal difference~\citep{munos2016safe, de2018multi, hernandez2019understanding}, or Monte Carlo returns~\citep{wright2013exploiting, wilcox2022monte}. To the best of our knowledge, these have been extensively studied in off-policy settings but are yet to be studied in offline settings. Also, Q-learning (Dynamic Programming) assumes the reward function to hold the Markov property, i.e. it must be a function of the current state and action while \ac{DT} and \ac{QDT} do not require such property. 
}
\pagebreak

{\bf Reproducing results for benchmarking.}
There have been many attempts to establish a benchmark for the offline \ac{RL} approaches by building datasets~\citep{fu2020d4rl,agarwal2020optimistic}, sharing their source code, as well as producing a library focusing on offline RL~\citep{seno2021d3rlpy}. However, we still found some conflicting results between papers. The leading cause of the issue is the requiring a vast amount of effort and computational power to reproduce the other researcher's results. As a result, most authors have no choice but to re-use the original results from state-of-the-art papers in the literature to establish a comparison. However, this leads to conflicting results due to the difficulties of reproducing all the details involved in these very diverse experimental setups. For example, many offline RL papers use D4RL MuJoCo datasets to evaluate their algorithms and compare them against other approaches. In this case, the datasets have three versions -- namely, v0, v1 and v2. While not always clearly stated, most papers use version v0. However, some use version v2, which causes some of the conflicting results. For example, \citet{chen2021decision} appears to evaluate their model with the v2 dataset while referencing other papers' results that use v0. A second issue with benchmarking the results in this manner is the usual insufficient number of simulations. As the simulations require large processing power, it is not feasible to run a large number of simulations. Most authors evaluate only 3 random seeds, which is often insufficient to compare the results. In this paper, we emphasise and analyse carefully the results from the simple environment, as it helps demonstrate the characteristics of the algorithm. Complex environments are still helpful; however, the estimated variance suggest that results should be handled with care when extracting conclusions.

\section{Conclusions}
\label{sec:conclusion}
We proposed Q-learning Decision Transformers, bringing the benefits of Dynamic Programming (Q-learning) approaches to reward conditioning sequence modelling methods to address some of their well-known weaknesses. Our approach provides a novel framework for improving offline reinforcement learning algorithms. In this paper, to illustrate the approach, we use existing state-of-the-art algorithms for both Dynamic Programming (\ac{CQL}) and reward conditioning modelling (\ac{DT}). Our evaluation shows the benefits of our approach over existing offline algorithms in line with the expected behaviour. Although the results are encouraging, there is room for improvement. For example, the \ac{QDT} results for Maze2D (Table~\ref{tab:maze2d_results}) are better than \ac{DT} but still not as good as \ac{CQL}. On the other hand, the \ac{QDT} results for Gym MuJoCo delayed reward (Table~\ref{tab:gym_mujoco_sparse_results}) are significantly better than \ac{CQL} but not as good as \ac{DT} in the walker2d. As we show in the discussion section, we can resolve these issues in certain situations. However, we further work is needed to resolve them for all environments.

{
{\bf Possible negative societal impact.}
Reinforcement learning algorithms (such as QDT) have a risk of being applied to potentially controversial fields with high impact in human lives -- e.g. military applications. These issues are inherited in all works in improving any autonomous systems.
}

{
\subsubsection*{Acknowledgments}
This work is supported by the UKRI Turing AI Fellowship EP/V024817/1 and SPHERE Next Steps Project funded by the U.K. Engineering, and Physical Sciences Research Council (EPSRC) under Grant EP/R005273/1.
}

\bibliography{library}
\bibliographystyle{icml2023}

\newpage
\appendix
\onecolumn

\section{Simple environment example trajectory data and its computation}
\label{apd:simple_env_example}
\begin{figure}[h]
    \hspace*{1.5cm}\includegraphics[width=0.9\textwidth]{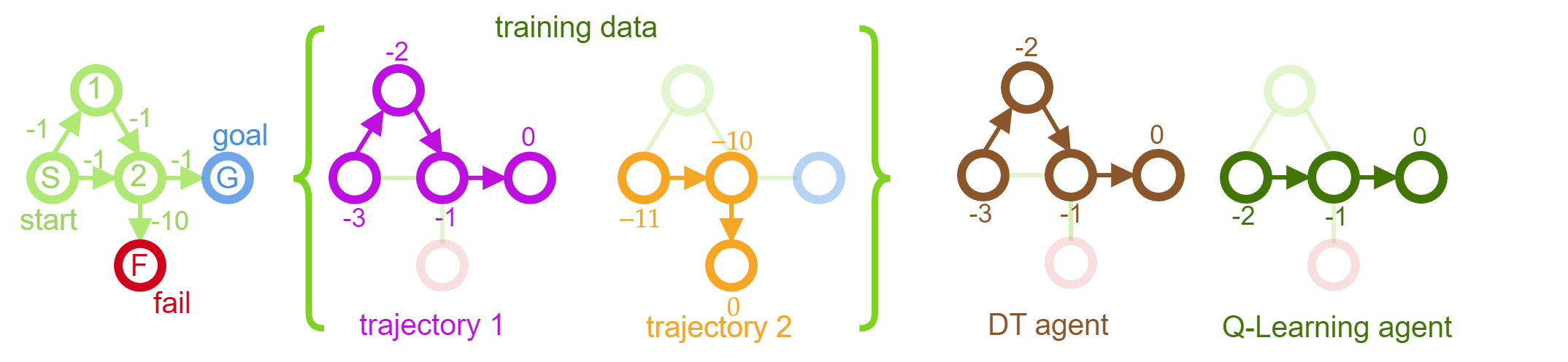}
    \caption{A simple example demonstrates the decision transformer approach's issue (lack of \textit{stitching} ability) -- fails to find the shortest path to the goal. In contrast, Q-learning finds the shortest path.}
    \label{fig:simple_example_apdx}
    \centering
\end{figure}

This section describes the trajectory data and some computation details for the simple example shown in Fig.~\ref{fig:simple_example_1}. We bring the figure here and added the state IDs in the circle (Fig.~\ref{fig:simple_example_apdx}).
The two trajectories of training data are: 
\begin{equation}
\begin{split}
trajectory 1 = [&s_0{=}S, a_0{=}up, r_0{=}{-}1, \\
&s_1{=}1, a_1{=}down, r_1{=}{-}1, \\
&s_2{=}2, a_2{=}right, r_2{=}{-}1, \\
&s_3{=}G, a_3{=}\text{N/A}, r_3{=}0] \\
trajectory 2 = [&s_0{=}S, a_0{=}right, r_0{=}{-}11, \\
&s_1{=}2, a_1{=}down, r_1{=}{-}10, \\
&s_2{=}F, a_2{=}\text{N/A}, r_2{=}0].
\end{split}
\end{equation}

We compute the \acf{RTG} from the reward $r_t$ as Eq.~\ref{eq:rtg-computation}. 
\begin{equation}
\label{eq:rtg-computation}
R_t = \sum_{\tau=0}^{T} r_\tau,
\end{equation}
where $R_t$ is \ac{RTG} at time step $t$ and $T$ is the episode length. The trajectories with the \acp{RTG} becomes as follows:
\begin{equation}
\begin{split}
trajectory 1 = [&s_0{=}S, a_0{=}up, r_0{=}{-}1, R_0{=}{-}3, \\
&s_1{=}1, a_1{=}down, r_1{=}{-}1,  R_1{=}{-}2,\\
&s_2{=}2, a_2{=}right, r_2{=}{-}1,  R_2{=}{-}1,\\
&s_3{=}G, a_3{=}\text{N/A}, r_3{=}0, R_3{=}0] \\
trajectory 2 = [&s_0{=}S, a_0{=}right, r_0{=}{-}1,  R_0{=}{-}11,\\
&s_1{=}2, a_1{=}down, r_1{=}{-}10, R_1{=}{-}10,\\
&s_2{=}F, a_2{=}\text{N/A}, r_2{=}0, R_2{=}0].
\end{split}
\end{equation}

\ac{DT} (the reward-conditioned approach) is trained to predict actions from the state and \ac{RTG}, so it takes $[s_t, R_t]$ as the input and outputs $a_t$. (Here, we assume the context length $K=1$ for \ac{DT} for simplicity.) For example, in the $t=0$ case, the \ac{DT} agent is trained to predict $a=up$ from $[s{=}S, R{=}{-}3]$ (trajectory 1) and $a=right$ from $[s{=}S, R{=}{-}11]$ (trajectory 2). For the evaluation, we set the \ac{RTG} the best value ($-2$ in this case) at $t=0$, and then the agent predicts the action from $[s{=}S, R{=}{-}2]$. Because the input $[s{=}S, R{=}{-}2]$ is closer to $[s{=}S, R{=}{-}3]$ (trajectory 1) than $[s{=}S, R{=}{-}11]$ (trajectory 2), the agent predict $a=up$ (trajectory 1) despite the optimal action is $a=right$ (trajectory 2). 

\section{Simple environment evaluation details}
\label{apd:simple_env_eval}
\subsection{Environment}
\begin{figure}[h]
    \centering
    \includegraphics[width=0.8\columnwidth]{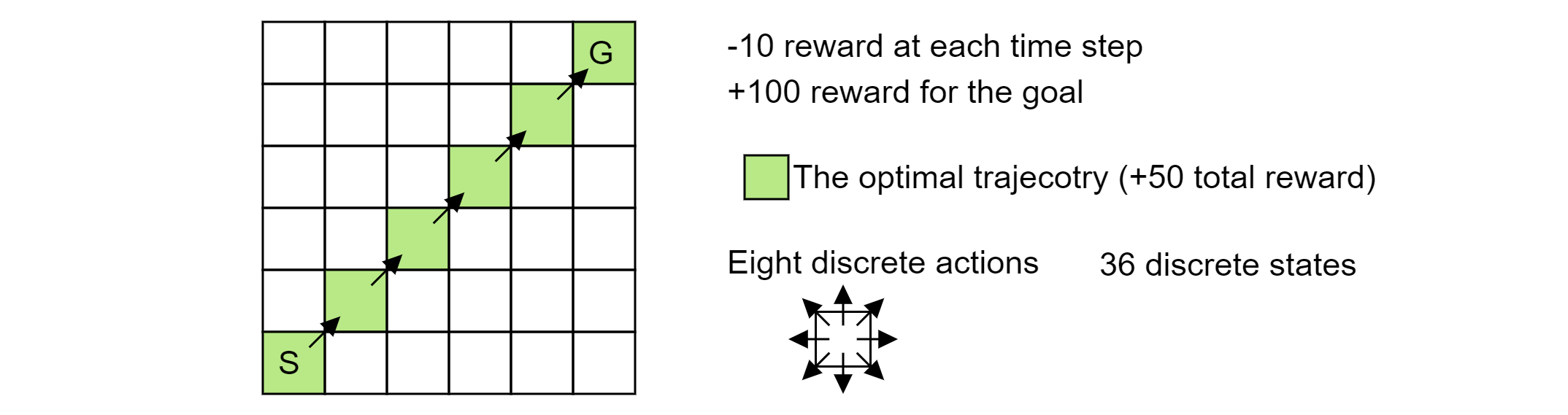}
    \caption{A simple 2D maze environment, which has 6-by-6 grid world and eight actions for moving eight directions. -10 reward at each time step and +100 reward for the goal. The optimal trajectory keeps moving up-right to the goal, which has total reward +50 ($=100-10*5$). The action is remapped so that the same action index is not always the optimal action. The mapping differs for each state, but fixed across the episodes.}
    \label{fig:simple_env}
\end{figure}

The environment has 6-by-6 discrete states and eight discrete actions as shown in Fig.~\ref{fig:simple_env}. The goal of the task is to find the shortest path from the start to the goal state. Each time step gives -10 reward and +100  reward at the goal. The optimal policy gives +50 total reward ($=100-10*5$). We also remap the action so that the same action index is not always optimal. The mapping differs for each state but is fixed across the episodes.

\subsection{Dataset}
We prepare an offline \ac{RL} dataset with a hundred episodes from a uniformly random policy and then remove an episode that achieves a positive total reward to make sure it only contains sub-optimal trajectories. As a result, the dataset used in this evaluation has one hundred episodes and 4,454 time steps. The maximum return of the hundred episodes is -10.0, the minimum return is -490 as we terminate the episode at 50 time step, and the average return is -415.5.

\subsection{CQL model details}
We build the CQL model for the simple environment based on Double Q-learning~\citep{Hasselt2010} and employ an embedding lookup table module to convert the discrete state to continuous high dimensional embedding space. The detailed model parameters are in Table~\ref{tab:simple-cql-model}.

\begin{table}[h]
\renewcommand{\arraystretch}{1.3}
  \centering
  \caption{Simple Enviornment CQL Model Parameters}
  \label{tab:simple-cql-model}
  \begin{tabular}{ll}
    \hline
    Parameter & Value \\
    \hline
    State embedding dimension   & 32 \\
    DQN type                    & fully connected \\
    DQN number of layers        & 2 \\
    DQN number of units         & 32 \\
    Optimizer                   & Adam \\
    Optimizer betas             & 0.9, 0.999 \\
    Optimizer learning rate     & 5.0e-4 \\
    Target network update rate  & 1.0e-2 \\
    Batch size                  & 128 \\
    Number of training steps    & 1000 updates \\
    Conservative weight ($\alpha$) & 0.5 \\
    \hline
  \end{tabular}
\end{table}

\subsection{DT and QDT model details}
Our DT and QDT model for the simple environment is constructed based on minGPT open-source code\footnote{https://github.com/karpathy/minGPT}. The detailed model parameters are in Table~\ref{tab:simple-dt-model}.

\begin{table}[h]
\renewcommand{\arraystretch}{1.3}
\centering
  \caption{Simple Environment DT/QDT Model Parameters}
  \label{tab:simple-dt-model}
  \begin{tabular}{ll}
    \hline
    Parameter & Value \\
    \hline
    Number of layers            & 4 \\
    Number of attention heads   & 4 \\
    Embedding dimension         & 64 \\
    Nonlinearity function       & ReLU \\
    Batch size                  & 64 \\
    Context length $K$          & 2 \\
    return-to-go conditioning   & 50 \\
    Dropout                     & 0.1 \\
    Learning rate               & 4.0e-4 \\
    \hline
  \end{tabular}
\end{table}

\subsection{Further evaluation results for simple environment}
The following tables have the simple environment results for all ten seeds. Table~\ref{tab:simple-full-results-1} shows the reward for the highest value during the training period. Table~\ref{tab:simple-full-results-2} shows the reward with the model at the end of training. 
DT and QDT have more significant differences between these two tables than the CQL results, which indicates that DT and QDT have overfitting issues and unstable learning behaviour.

\begin{table}[h]
\renewcommand{\arraystretch}{1.3}
  \centering
  \caption{Simple Environment Full Results (Best). The results from the best performing model during the training.}
  \label{tab:simple-full-results-1}
  \begin{tabular}{p{0.8cm}ccc}
    \hline
    & CQL & DT & QDT \\
    \hline
    \multirow{10}*{\rotatebox{90}{\parbox{2.5cm}{\centering{results for ten random seeds}}}} &
          40.0 & 18.2 & 43.6 \\
        & 40.0 & 20.4 & 42.0 \\
        & 40.0 & 11.2 & 49.2 \\
        & 40.0 & 13.8 & 42.6 \\
        & 40.0 & 12.6 & 39.2 \\
        & 40.0 &  8.4 & 27.8 \\
        & 40.0 & 19.6 & 47.2 \\
        & 40.0 & 21.2 & 47.4 \\
        & 40.0 & 14.4 & 37.4 \\
        & 40.0 & 18.8 & 46.0 \\
    \hline
    mean & 40.0 & 15.9 & 42.2 \\
    std. & 0.0  & 4.4  & 6.3 \\
    \hline
  \end{tabular}
\end{table}

\begin{table}[h]
\renewcommand{\arraystretch}{1.3}
  \centering
  \caption{Simple Environment Full Results (Last). The results from the model at the end of the training.}
  \label{tab:simple-full-results-2}
  \begin{tabular}{p{0.8cm}ccc}
    \hline
    & CQL & DT & QDT \\
    \hline
    \multirow{10}*{\rotatebox{90}{\parbox{2.5cm}{\centering{results for ten random seeds}}}} &
          40.0 & -39.2 & 13.8 \\
        & 40.0 &   8.6 & 35.8 \\
        & 40.0 & -25.4 & 46.6 \\
        & 40.0 & -20.8 & 16.6 \\
        & 30.0 & -50.2 & 29.2 \\
        & 40.0 & -26.0 & 19.6 \\
        & 40.0 &   9.4 & 44.0 \\
        & 30.0 & -35.0 & 47.4 \\
        & 40.0 & -10.2 & 23.2 \\
        & 40.0 &   7.8 & 35.0 \\
    \hline
    mean & 38.0 & -18.1 & 31.1 \\
    std. & 4.2  & 21.3  & 12.5 \\
    \hline
  \end{tabular}
\end{table}

\section{Open AI Gym MuJoCo and Maze2D evaluation details}
\subsection{CQL model details}
For MuJoCo Gym CQL evaluation, we use d3rlpy library~\citep{seno2021d3rlpy}. It provides a script to run the evaluation (d3rlpy/reproduce/offline/cql.py), and it uses the same hyperparameters as \citet{kumar2020conservative}.
For Mazed2d simulations, we re-use the same d3rlpy script with the same hyperparameter settings.

\subsection{DT and QDT model details}
For DT simulations, we use the code provided by the original paper authros\footnote{https://github.com/kzl/decision-transformer} for both MuJoCo Gym and Maze2D environments. For QDT simulations, we added extra code to relabelling the return-to-go to the DT script (decision-transformer/gym/experiment.py). The relabelling code is described in Algorithm~\ref{alg:replace_R} and \ref{alg:replace_R_2}.

\subsection{Evaluation Process}
\paragraph{CQL} We train the {CQL} model with five random seeds for 500,000 updates with 256 batch size, then evaluate the model at the end of the training with 10 episode roll-outs. We inherit these \ac{CQL} settings from d3rlpy offline \ac{RL} library~\citep{seno2021d3rlpy}.

\paragraph{DT} We train the \ac{DT} model with five random seeds for 100,000 updates with 64 batch size, then evaluate the model at the end of the training with 100 episode roll-outs. We inherit these \ac{DT} settings from the source code provided by the \ac{DT} paper authors\footnote{https://github.com/kzl/decision-transformer}~\citep{chen2021decision}.

\paragraph{QDT} We train the \ac{QDT} model with five random seeds, each of them employing its own trained \ac{CQL} model to relabel the dataset. \ac{QDT} model is trained for 100,000 updates for MuJoCo Gym and 150,000 updates for maze2d with 64 batch size, then evaluate the model at the end of the training with 100 episode roll-outs -- same as \ac{DT}. 

\subsection{Hyper parameter search}
We use the same hyper-parameter settings as the original papers~\citep{kumar2020conservative, chen2021decision}. However, we did some hyper-parameter searches for the conservative weight ($\alpha$). It is because the optimal conservative weight value could be different for CQL and QDT.  

For MuJoCo Gym environments, we start with $\alpha=10.0$ for medium dataset and $\alpha=5.0$ for medium-replay dataset. We take these values from the \ac{CQL} paper. Then, reduce these values to see if the performance of \ac{CQL} and \ac{QDT} varies. Table~\ref{tab:parameter_sweep_cql_gym} and Table~\ref{tab:parameter_sweep_qdt_gym} shows \ac{CQL} and \ac{QDT} results respectively.
These results show that $\alpha=10.0$ for medium dataset and $\alpha=5.0$ for medium-replay dataset perform well for QDT and do not degrade performance significantly for CQL. Also, they are the same values as the original paper, so we decide to keep them the same as the paper.

\begin{table*}[h]
  \centering
    \caption{CQL results for Open AI Gym MuJoCo with conservative weight parameter ($\alpha$) sweep. Average and standard deviation scores are reported over three seeds.}
    \label{tab:parameter_sweep_cql_gym}
    \begin{tabular}{>{\centering\arraybackslash}p{0.5cm}
                  p{2.3cm}
                  >{\centering\arraybackslash}p{2.0cm}
                  >{\centering\arraybackslash}p{2.0cm}
                  >{\centering\arraybackslash}p{2.0cm}
                  >{\centering\arraybackslash}p{2.0cm}
                  }
    \hline
        &         & \multicolumn{4}{c}{CQL} \\
        & Dataset & $\alpha=10.0$ & $\alpha=5.0$ & $\alpha=2.5$  & $\alpha=1.25$ \\
    \hline
    \multirow{3}*{\RotText{Medium}} 
        & Hopper-v2      & $68.7\pm16.4$ & $72.5\pm9.5$          & $\mathbf{83.6}\pm3.8$  & \rule[0mm]{0mm}{3.2mm} \\
        & HalfCheetah-v2 & $48.9\pm2.4$  & $51.8\pm2.4$          & $\mathbf{57.0}\pm1.1$  & \rule[0mm]{0mm}{3.2mm} \\
        & Walker2d-v2    & $83.3\pm0.5$  & $\mathbf{86.2}\pm0.5$ & $43.5\pm43.6$ & \rule[0mm]{0mm}{3.2mm} \\
    \hline
    \multirow{3}*{\RotText{Medium Replay}}
        & Hopper-v2      & & $\mathbf{95.4}\pm11.6$ & $87.5\pm24.7$ & $90.7\pm14.5$ \rule[0mm]{0mm}{3.2mm}\\
        & HalfCheetah-v2 & & $49.9\pm2.9$           & $51.8\pm2.7$  & $\mathbf{54.3}\pm0.2$  \rule[0mm]{0mm}{3.2mm}\\
        & Walker2d-v2    & & $\mathbf{88.9}\pm3.7$  & $50.6\pm36.3$ & $16.8\pm14.2$ \rule[0mm]{0mm}{3.2mm}\\
    \hline
  \end{tabular}
\end{table*}

\begin{table*}[h]
  \centering
    \caption{QDT results for Open AI Gym MuJoCo with conservative weight parameter ($\alpha$) sweep. Average and standard deviation scores are reported over three seeds.}
    \label{tab:parameter_sweep_qdt_gym}
    \begin{tabular}{>{\centering\arraybackslash}p{0.5cm}
                  p{2.3cm}
                  >{\centering\arraybackslash}p{2.0cm}
                  >{\centering\arraybackslash}p{2.0cm}
                  >{\centering\arraybackslash}p{2.0cm}
                  >{\centering\arraybackslash}p{2.0cm}
                  }
    \hline
        &         & \multicolumn{4}{c}{QDT} \\
        & Dataset & $\alpha=10.0$ & $\alpha=5.0$ & $\alpha=2.5$  & $\alpha=1.25$ \\
    \hline
    \multirow{3}*{\RotText{Medium}} 
        & Hopper-v2      & $\mathbf{68.6}\pm7.5$ & $65.3\pm1.3$          & $57.5\pm6.6$ & \rule[0mm]{0mm}{3.2mm} \\
        & HalfCheetah-v2 & $\mathbf{42.2}\pm0.5$ & $\mathbf{42.2}\pm0.05$& $42.1\pm0.4$ & \rule[0mm]{0mm}{3.2mm} \\
        & Walker2d-v2    & $65.9\pm3.6$          & $\mathbf{70.1}\pm2.4$ & $68.8\pm6.9$ & \rule[0mm]{0mm}{3.2mm} \\
    \hline
    \multirow{3}*{\RotText{Medium Replay}}
        & Hopper-v2      & & $55.3\pm28.0$         & $40.2\pm5.9$          & $\mathbf{64.0}\pm22.9$ \rule[0mm]{0mm}{3.2mm}\\
        & HalfCheetah-v2 & & $\mathbf{35.7}\pm0.6$ & $35.5\pm0.4$          & $33.0\pm0.5$  \rule[0mm]{0mm}{3.2mm}\\
        & Walker2d-v2    & & $59.1\pm2.8$          & $\mathbf{64.3}\pm5.9$ & $45.2\pm39.5$ \rule[0mm]{0mm}{3.2mm}\\
    \hline
  \end{tabular}
\end{table*}

For maze2d environment, we start with $\alpha=10.0$ which is the value used in the CQL paper for MuJoCo Gym environments medium datasets. Then, reducing these values to see if the performance of \ac{CQL} varies. Table~\ref{tab:parameter_sweep_cql_maze2d} shows the simulation results. We pick $\alpha=1.0$ as it performs the best. It is possible that even lower values might perform better. We see \ac{QDT} shows good improvement over \ac{DT} with $\alpha=1.0$, so we use the value for this paper. We would like to try further optimisation in the future.

\begin{table*}[h]
  \centering
    \caption{CQL results for Maze2D with conservative weight parameter ($\alpha$) sweep. Average and standard deviation scores are reported over three seeds.}
    \label{tab:parameter_sweep_cql_maze2d}
    \begin{tabular}{
                  p{3cm}
                  >{\centering\arraybackslash}p{2.0cm}
                  >{\centering\arraybackslash}p{2.0cm}
                  >{\centering\arraybackslash}p{2.0cm}
                  }
    \hline
                & \multicolumn{3}{c}{CQL} \\
        Dataset & $\alpha=10.0$ & $\alpha=2.0$ & $\alpha=1.0$ \\
    \hline
        maze2d-umaze-v1  & $27.3\pm12.2$ & $66.1\pm9.8$ & $\mathbf{96.0}\pm32.2$ \rule[0mm]{0mm}{3.2mm} \\
        maze2d-medium-v1 & $-3.5\pm1.3$  & $\mathbf{36.6}\pm3.7$ & $35.9\pm15.3$ \rule[0mm]{0mm}{3.2mm} \\
        maze2d-large-v1  & $-2.5\pm0.0$  & $40.8\pm6.0$ & $\mathbf{53.2}\pm7.0$ \rule[0mm]{0mm}{3.2mm} \\
    \hline
  \end{tabular}
\end{table*}

\section{Justification of replacing RTG with the learned value function}
\label{apd:relabelling_rtg}
Define the optimal state value function as $V^*(s_t)$, the learned lower bound of the value function as $\hat{V}(s_t)$ and the corresponding return-to-go value as $R_t$. We show that when $\hat{V}(s_t)>R_t$, the error in $\hat{V}(s_t)$ is smaller than the error in $R_t$. We start from the condition,
\begin{equation}
\begin{split}
    \hat{V}(s_t) &> R_t \\
    V^*(s_t) - \hat{V}(s_t) &< V^*(s_t) - R_t.
\end{split}
\end{equation}
As $\hat{V}(s_t)$ is the lower bound of $V^*(s_t)$, $V^*(s_t) \geq \hat{V}(s_t)$. Hence both sides of the above equation are non-negative. We can take the absolute of both terms, and we get,
\begin{equation}
    |V^*(s_t) - \hat{V}(s_t)| < |V^*(s_t) - R_t|.
\end{equation}
This indicates that the error in $\hat{V}(s_t)$ is smaller than the error in $R_t$.

\section{Further Discussions}

\subsection{Why CQL outperforms DT/QDT on Maze2D, but fails on MuJoCo Gym delayed reward?}
It is because maze2d are simpler environments and have shorter episodes than the MuJoCo control tasks.
Table~\ref{tab:gym_maze2d_comparison} shows that the action dimension, the state (observation) dimension and the episode length averaged over the top 5\% returns in the dataset. It can be seen that MuJoCo tasks have higher action/state dimensions and longer episode lengths than Maze2d. Also, the evaluation results for the Sparse maze2d-medium and -large show some notable performance loss against the Dense counterparts, which is aligned with the fact that their episode lengths are longer than the maze2d-open and -umaze.

\begin{table*}[h]
  \centering
    \caption{MuJoCo Gym and Maze2D environments comparison. The table shows that the action dimension, the state (observation) dimension and the episode length averaged over the top 5\% returns in the dataset.}
    \label{tab:gym_maze2d_comparison}
    \begin{tabular}{
                  p{3cm}
                  >{\centering\arraybackslash}p{2.0cm}
                  >{\centering\arraybackslash}p{2.0cm}
                  >{\centering\arraybackslash}p{3.0cm}
                  }
    \hline
        Environment & Action Dimension & State Dimension & Good Episode Average Length \\
    \hline
        hopper      & 3 & 11 & 708.2  \rule[0mm]{0mm}{3.2mm} \\
        halfcheetah & 6 & 17 & 1000.0 \rule[0mm]{0mm}{3.2mm} \\
        walker2d    & 6 & 17 & 996.7  \rule[0mm]{0mm}{3.2mm} \\
    \hline
        maze2d-open   & 2 & 4 & 49.8 \rule[0mm]{0mm}{3.2mm} \\
        maze2d-umanze & 2 & 4 & 128.6 \rule[0mm]{0mm}{3.2mm} \\
        maze2d-medium & 2 & 4 & 224.1 \rule[0mm]{0mm}{3.2mm} \\
        maze2d-large  & 2 & 4 & 314.6 \rule[0mm]{0mm}{3.2mm} \\
    \hline
  \end{tabular}
\end{table*}

\subsection{Why QDT outperforms DT on Maze2D whereas it does not on Gym despite both having dense rewards?}

It is due to the difference in the training data. maze2d dataset is designed to test the stitching ability; hence it only has sub-optimal trajectories, whereas the MuJoCo Gym dataset has some optimal trajectories. If the dataset has some optimal trajectories, DT will perform well. On the other hand, if the dataset has only suboptimal trajectories, DT will struggle, and QDT improves such cases by utilising the information in CQL.

As maze2d only has suboptimal trajectories, DT struggles with them, and QDT can perform better than DT. For MuJoCo Gym cases, the dataset has some optimal trajectories; hence DT performs well, and so as QDT.

Strictly speaking, there are some exceptions. MuJoCo halfcheetah-medium and halfcheetah-medium-replay dataset does not have an optimal trajectory, still QDT performs similarly to DT. It is because even CQL struggles to achieve good performance on these datasets. (CQL only performs similarly to DT even though CQL can stitch the suboptimal trajectories.) As CQL struggled, QDT could not get much help from CQL.

The other exception is maze2d-open and maze2d-open-dense. These datasets have good trajectories. It is actually aligned with our evaluation results. The results for maze2d-open and maze2d-open-dense show good performance with DT.

Table~\ref{tab:gym_maze2d_dataset_comparison} shows the maximum, 95 percentile and 90 percentile values of the normalised returns (score) in the dataset. As we discussed above, Maze2d has suboptimal trajectories (except open and open-dense), and MuJoCo Gym has (near) optimal trajectories -- a score close to 100 (except halfcheetah).

\begin{table*}[h]
  \centering
    \caption{Scores in MuJoCo Gym and Maze2D datasets. This table shows that maximum score, 95 percentile score and 90 percentile score values for each dataset.}
    \label{tab:gym_maze2d_dataset_comparison}
    \begin{tabular}{
                  p{4.5cm}
                  >{\centering\arraybackslash}p{2.0cm}
                  >{\centering\arraybackslash}p{2.0cm}
                  >{\centering\arraybackslash}p{2.0cm}
                  }
    \hline
        Dataset & max. score & 95 pct. score & 90 pct. score \\
    \hline
        maze2d-open-v0        & 232.4 & 130.7 & 116.2  \rule[0mm]{0mm}{3.2mm} \\
        maze2d-open-dense-v0  & 188.9 & 128.4 & 117.4  \rule[0mm]{0mm}{3.2mm} \\
        maze2d-umaze-v1       & 21.1  & 13.2  & 10.3   \rule[0mm]{0mm}{3.2mm} \\
        maze2d-umaze-dense-v1 & -1.4 & -11.7 & -18.3  \rule[0mm]{0mm}{3.2mm} \\
        maze2d-medium-v1      & 12.8  & 6.8   & 4.9    \rule[0mm]{0mm}{3.2mm} \\
        maze2d-medium-dense-v1&8.9  & 4.0   & 0.3    \rule[0mm]{0mm}{3.2mm} \\
        maze2d-large-v1       & 16.9  & 6.5   & -2.5   \rule[0mm]{0mm}{3.2mm} \\
        maze2d-large-dense-v1 & 14.6& 7.9   & -2.4   \rule[0mm]{0mm}{3.2mm} \\
    \hline
        hopper-medium-v2        & 99.5 & 63.2 & 57.0 \rule[0mm]{0mm}{3.2mm} \\
        hopper-medium-replay-v2 & 98.6 & 46.4 & 31.5 \rule[0mm]{0mm}{3.2mm} \\
        halfcheetah-medium-v2   & 45.0 & 43.0 & 42.5 \rule[0mm]{0mm}{3.2mm} \\
        halfcheetah-medium-replay-v2 &42.4 & 39.9 & 39.2 \rule[0mm]{0mm}{3.2mm} \\
        walker2d-medium-v2      & 92.0 & 83.4 & 82.4 \rule[0mm]{0mm}{3.2mm} \\
        walker2d-medium-replay-v2 & 89.9 & 66.6 & 42.5 \rule[0mm]{0mm}{3.2mm} \\
    \hline
  \end{tabular}
\end{table*}

\subsection{Why QDT performs close to DT, not CQL in Fig.~\ref{fig:x-pct-hopper} (Gym hopper)?}

The main reason is that QDT employs DT as its agent algorithm. The difference lays in its training data. 
If the environment/dataset has specific characteristics that work against DT approach, those also work against QDT. Some of these properties, such as dataset sub-optimality, are fixed/mitigated by QDT.
However, there may be other elements that are against DT and QDT, e.g., the environment having a few critical states~\citep{kumar2022should}. If this is also behind the gap between CQL and DT, then it is possible QDT performs close/the same as DT.

\citet{kumar2022should} studied the Dynamic Programming approach and the imitation learning approach and compared the upper bounds of their sub-optimality (the difference between the return
from the optimal policy and the learned policy). They show that the Dynamic Programming approach is preferred over imitation learning when the environment has a few critical states --  the return of the episode mostly depends upon the actions in these states. The results in \citet{kumar2022should} are based on theoretical analysis (sub-optimality upper bounds). Hence, it is possible that the imitation learning approach (DT and QDT) can perform as well as or better than Dynamic Programming approaches (such as CQL) in practice. \citet{kumar2022should} empirically shows that the goal-conditioned approach remains competitive by selecting the right level of model capacity and the goal. There are still many open and ongoing discussions regarding the comparison.

\subsection{Extra results for removing top X\%}

\begin{figure}[h]
\centering
\begin{minipage}{.5\textwidth}
  \centering
  \includegraphics[width=1.0\linewidth]{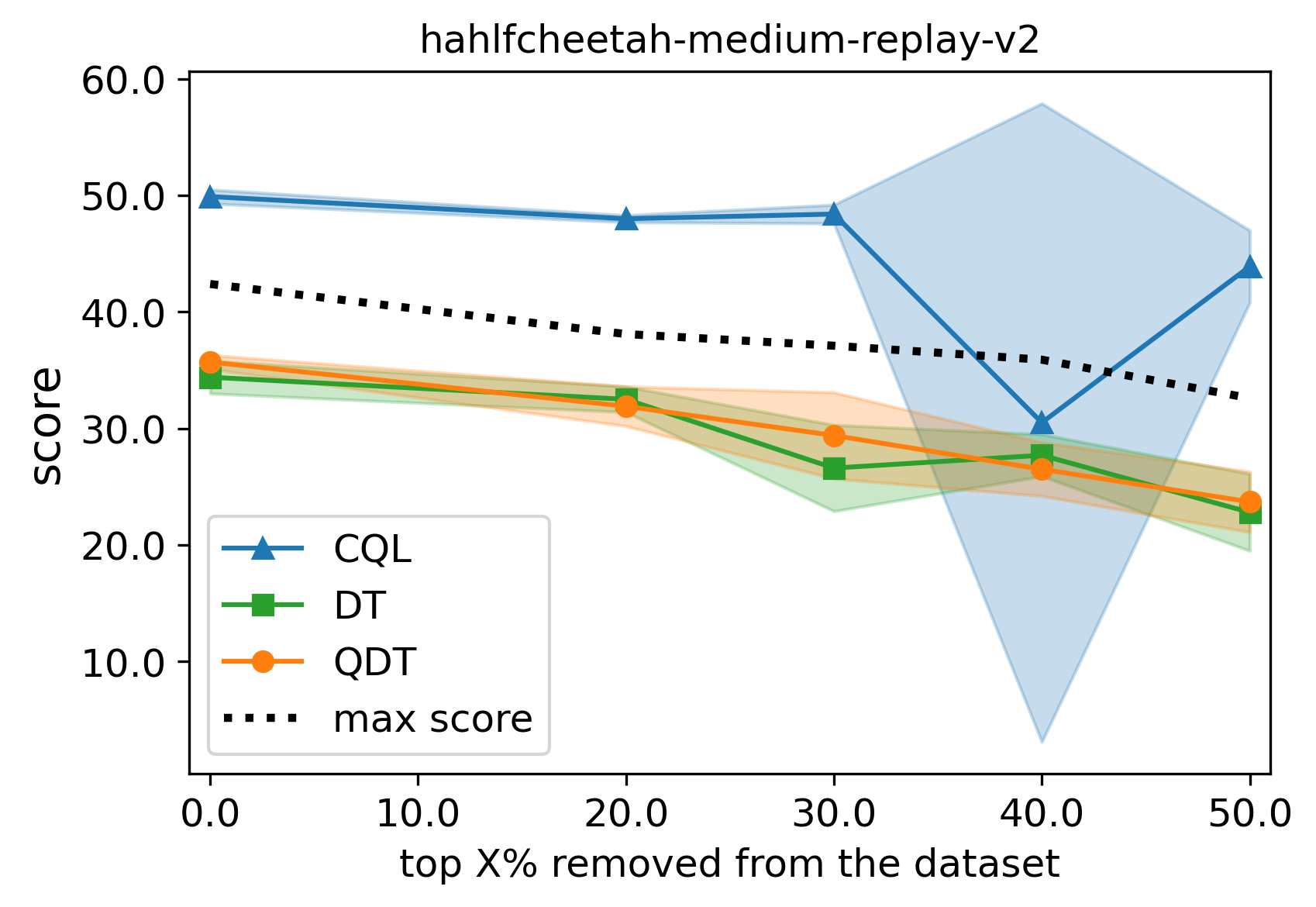}
\end{minipage}%
\begin{minipage}{.5\textwidth}
  \centering
  \includegraphics[width=1.0\linewidth]{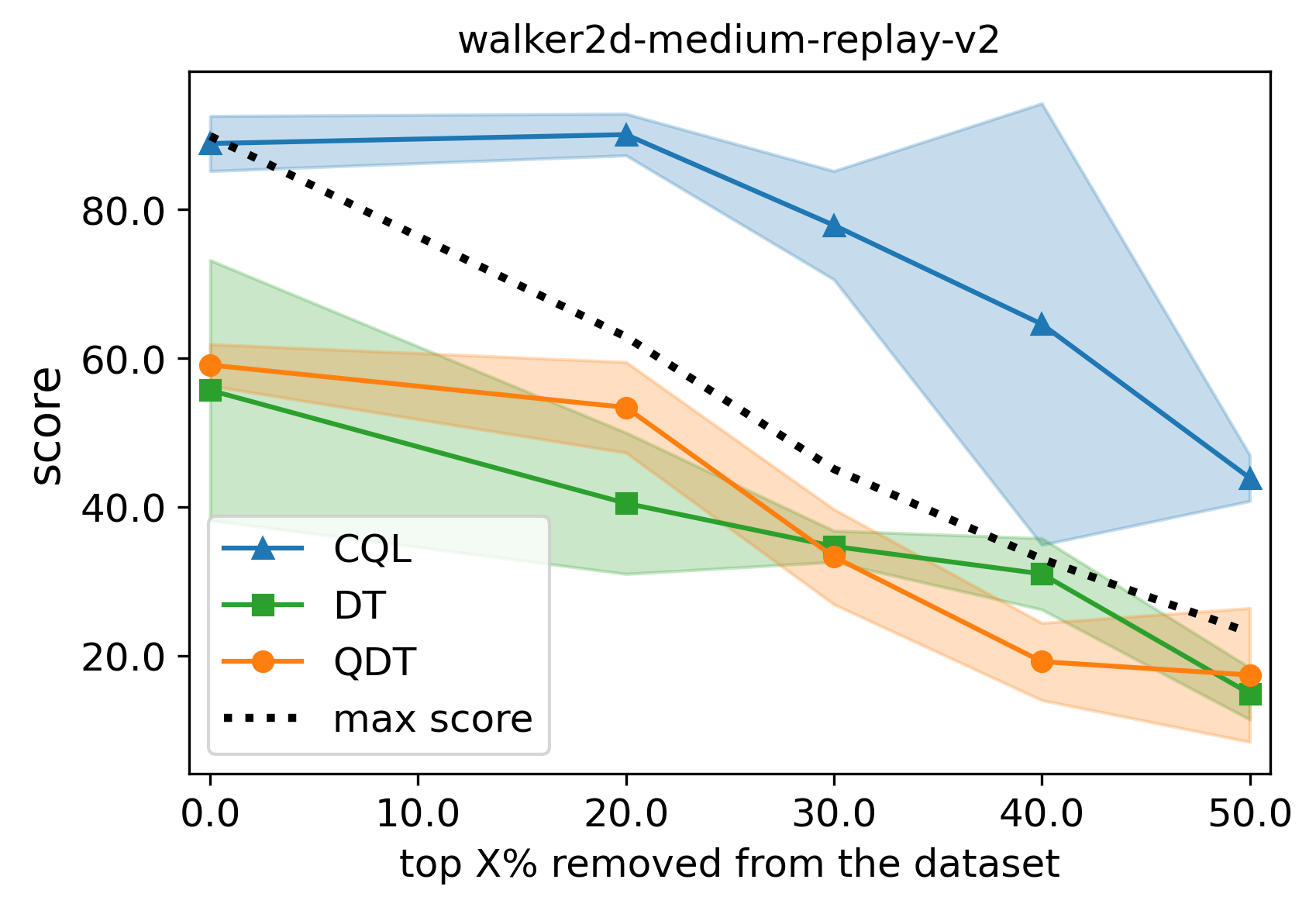}
\end{minipage}
\caption{Evaluation results (scores) for CQL, DT and QDT with the halfcheetah-medium-replay-v2 and walker2d-medium-replay-v2 dataset removed top X\% trajectories. The shaded area shows one standard deviation range of the results. It also has the maximum score in the dataset as a reference.}\label{fig:x-pct-halfcheetah-walker2d}
\end{figure}

We run the same experiment as the Stitching ability subsection in Section~\ref{sec:discussion} on the other two MuJoCo Gym environments. The results (Fig.~\ref{fig:x-pct-halfcheetah-walker2d}) do not show a clear benefit of QDT over DT. We think it is because the cause of the gap between CQL and DT is not just the sub-optimality in the dataset (still the sub-optimality can be the cause of the difference, but it is not the only cause in these cases.)

\subsection{Consistency relabelling ablation experiment}
We have tried the ablation experiment for the consistency relabelling (Algorithm 2) on a subset of environments. The results are summarised it in Table~\ref{tab:alg2_ablation_exmepriment}. We run ten random seeds for Simple Environment and three for others.
The Simple Environment result shows some benefits of using Algorithm 2 in its average value, although it is not significant. 
For the other more complex environments, we do not see clear benefits of Algorithm2. We think this is because the changes applied by Algorithm2 are relatively minor compared to the original RTG variations. We think it is better to keep Algorithm 2, at least for now because the training data could have non-realistic (inconsistent) samples without the algorithm. 

\begin{table*}[h]
  \centering
    \caption{Scores in Simple Environment, MuJoCo Gym and Maze2D datasets. This table shows QDT results and QDT without the consistency relabelling (Algorithm 2).}
    \label{tab:alg2_ablation_exmepriment}
    \begin{tabular}{
                  p{4.5cm}
                  >{\centering\arraybackslash}p{2.0cm}
                  >{\centering\arraybackslash}p{2.0cm}
                  }
    \hline
        Dataset & QDT & QDT w/o Alg.2 \\
    \hline
        Simple Environment  & $42.2\pm6.3$ & $29.7\pm13.8$ \rule[0mm]{0mm}{3.2mm} \\
    \hline
        hopper-medium-v2      & $65.3\pm2.0$ & $65.7\pm3.9$ \rule[0mm]{0mm}{3.2mm} \\
        halfcheetah-medium-v2 & $42.2\pm2.3$ & $42.4\pm0.1$ \rule[0mm]{0mm}{3.2mm} \\
        walker2d-medium-v2    & $71.3\pm2.4$ & $80.2\pm10.8$\rule[0mm]{0mm}{3.2mm} \\
    \hline
        maze2d-large-v1       & $35.0\pm24.2$& $23.0\pm5.0$ \rule[0mm]{0mm}{3.2mm} \\
    \hline
  \end{tabular}
\end{table*}

\subsection{Aggregated evaluation results}
\begin{figure}[h]
\centering
\centering
  \includegraphics[width=1.0\linewidth]{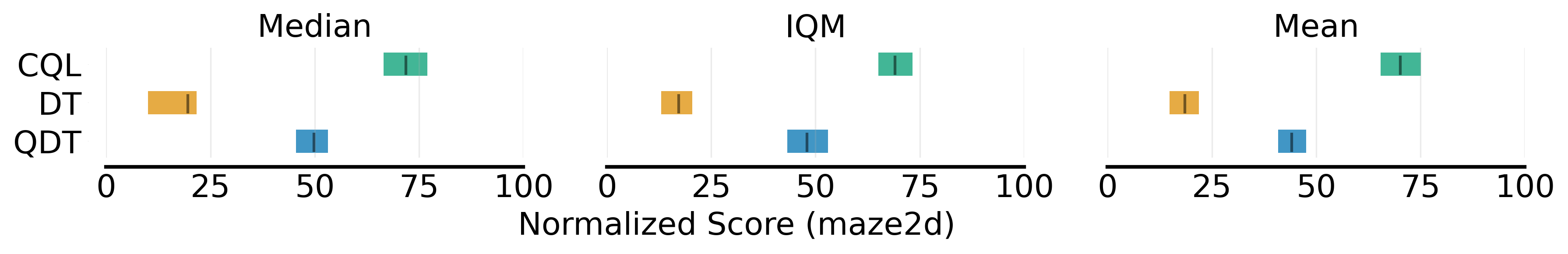}
  \includegraphics[width=1.0\linewidth]{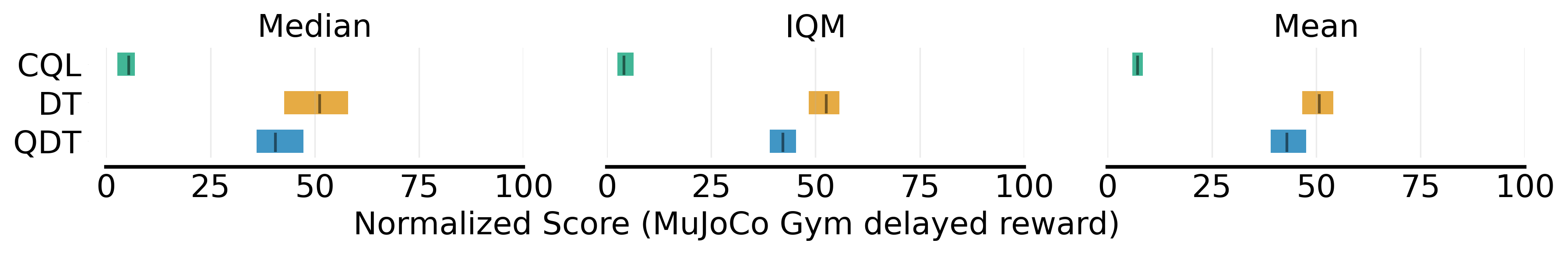}
  \includegraphics[width=1.0\linewidth]{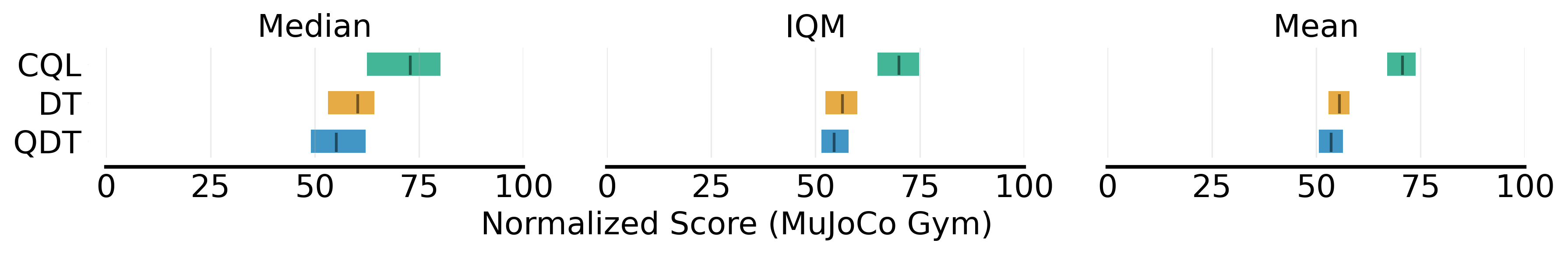}
\caption{Aggregated evaluation results (scores) for each group of environments (maze2d, MuJoCo Gym delayed reward and MuJoCo Gym) with three different metrics -- median, Interquantile mean (IQM) and mean. The results support our conclusions 1) DT struggles in maze2d, but QDT improves DT by getting help from CQL. 2) CQL fails in MuJoCo Gym delayed reward. 3) DT and QDT perform similarly in MuJoCo Gym.}\label{fig:results-CI}
\end{figure}

We compute the aggregated evaluation results for each group of environments (maze2d, MuJoCo Gym delayed reward and MuJoCo Gym) with three different metrics -- median, Interquantile mean (IQM) and mean (Fig.~\ref{fig:results-CI}). It uses 95\% stratified bootstrapped confidence interval~\citep{agarwal2021deep}.

The results support our conclusions 1) DT struggles in maze2d, but QDT improves DT performance by getting help from CQL. 2) CQL fails in MuJoCo Gym delayed reward. 3) DT and QDT perform similarly in MuJoCo Gym. Note that QDT has never failed in any of these groups of environments.

\acrodef{RL}{reinforcement learning}
\acrodef{UDRL}{upside down reinforcement learning}
\acrodef{DT}{decision transformer}
\acrodef{CQL}{conservative Q learning}
\acrodef{QDT}{Q-learning Decision Transformer}
\acrodef{BC}{behaviour cloning}
\acrodef{DRL}{deep reinforcement learning}
\acrodef{DQN}{deep Q-learning}
\acrodef{DNN}{deep neural network}
\acrodef{MDP}{Markov decision process}
\acrodef{RTG}{return-to-go}


\end{document}